\documentclass{article} % For LaTeX2e
\usepackage{iclr2024_conference,times}

\usepackage{amsmath,amsfonts,bm}

\def\eqref#1{equation~\ref{#1}}
\def\1{\bm{1}}

\DeclareMathAlphabet{\mathsfit}{\encodingdefault}{\sfdefault}{m}{sl}
\SetMathAlphabet{\mathsfit}{bold}{\encodingdefault}{\sfdefault}{bx}{n}

\usepackage{hyperref}
\usepackage{url}
\usepackage{graphicx}
\usepackage{booktabs} % for professional tables
\usepackage{hyperref}
\usepackage{subfig}
\usepackage[export]{adjustbox}
\usepackage{multirow}

\usepackage{amsmath}
\usepackage{amssymb}
\usepackage{mathtools}
\usepackage{amsthm}

\usepackage[capitalize,noabbrev]{cleveref}

\title{Memorization Through the Lens of Curvature of Loss Function Around Samples}

\author{Isha Garg\thanks{Corresponding Author}~, Deepak Ravikumar, Kaushik Roy\\
Department of Electrical and Computer Engineering, \\
Purdue University\\
West Lafayette, IN 47906, USA\\
\texttt{\{gargi,dravikum,kaushik\}@purdue.edu} \\
}

\iclrfinalcopy % Uncomment for camera-ready version, but NOT for submission.
\begin{document}

\maketitle

\begin{abstract}
Deep neural networks are over-parameterized and easily overfit the datasets they train on. In the extreme case, it has been shown that these networks can memorize a training set with fully randomized labels. We propose using the curvature of loss function around each training sample, averaged over training epochs, as a measure of memorization of the sample. We use this metric to study the generalization versus memorization properties of different samples in popular image datasets and show that it captures memorization statistics well, both qualitatively and quantitatively. We first show that the high curvature samples visually correspond to long-tailed, mislabeled, or conflicting samples, those that are most likely to be memorized. This analysis helps us find, to the best of our knowledge, a novel failure mode on the CIFAR100 and ImageNet datasets: that of duplicated images with differing labels. Quantitatively, we corroborate the validity of our scores via two methods. First, we validate our scores against an independent and comprehensively calculated baseline, by showing high cosine similarity with the memorization scores released by \citet{feldman}.  Second, we inject corrupted samples which are memorized by the network, and show that these are learned with high curvature. 
To this end, we synthetically mislabel a random subset of the dataset. We overfit a network to it and show that sorting by curvature yields high AUROC values for identifying the corrupted samples. An added advantage of our method is that it is scalable, as it requires training only a single network as opposed to the thousands trained by the baseline, while capturing the aforementioned failure mode that the baseline fails to identify.

% sorting by curvature yields high AUROC values for identifying the mislabeled samples.
% \vspace{-2mm}
\end{abstract}
% \vspace{-2mm}

\vspace{-2mm}
\section{Introduction}
\label{sec:intro}

Deep learning has been hugely successful in many fields. With increasing availability of data and computing capacity, networks are getting larger, growing to billions of parameters. This overparametrization often results in the problem of overfitting. An extreme form of overfitting was demonstrated by \citet{zhang2017understanding}, who showed that networks can memorize a training set with fully randomized labels. Further, networks make overconfident predictions, even when the predictions are incorrect \citep{calibration}, memorizing both mislabeled and long-tailed outliers alike. This can prove to be harmful in real-world settings, such as with data poisoning attacks\citep{datapoision2012, chen2017targeted} and privacy leakage in the form of membership inference attacks \citep{membership}.
% wherein attackers can find the exact samples used to train the network. 
There has been considerable research effort towards countering overfitting, with different forms of regularization \citep{wtdecay}, dropout \citep{dropout}, early stopping \citep{earlystop} and data augmentation \citep{cutout, mixup}. In this paper, we exploit overfitting to propose a new metric for measuring memorization of a data point, that of the curvature of the network loss around a data point. We overfit intentionally and utilize this to study the samples the network is memorizing.%, and how memorization progresses with training. 

\begin{figure*}[!tp]
  \centering
  \includegraphics[width=1.0\linewidth]{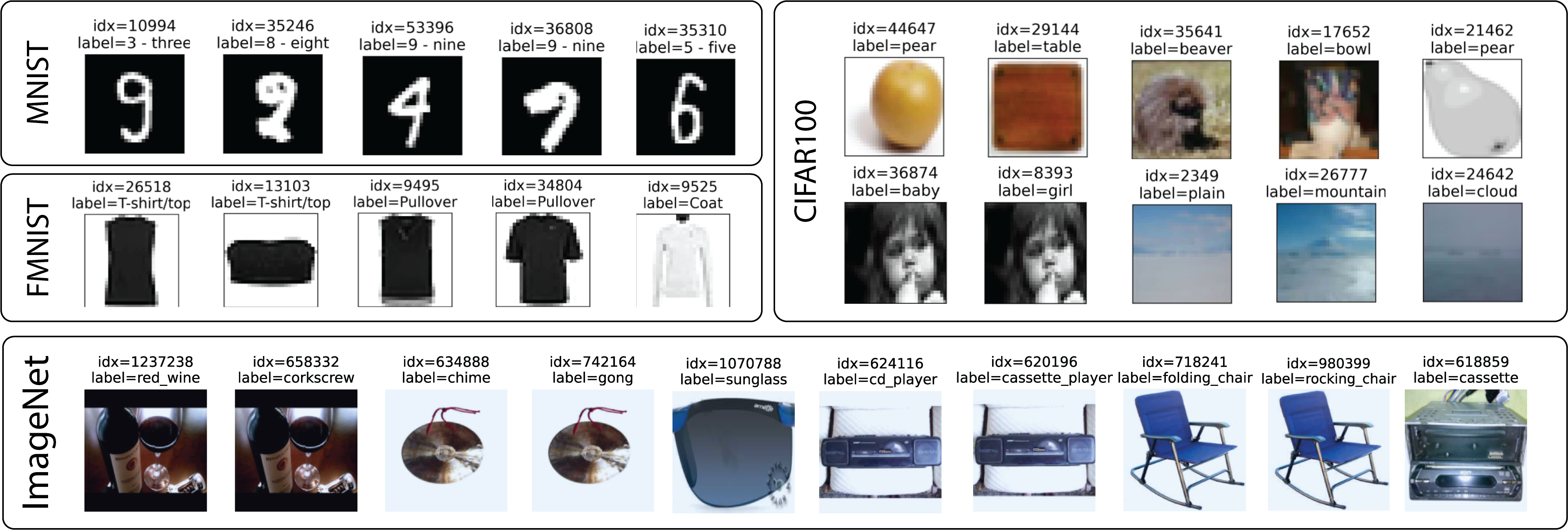}
  \caption{Cherry picked examples from the 100 highest curvature training examples on MNIST, FMNIST, CIFAR100 and ImageNet.}
  \vspace{-2mm}
  \label{fig:cherry}
  \end{figure*}

We focus our analysis on examples with excessive loss curvature around them compared to other training data samples. We find that such samples with high curvature are rare instances, i.e. drawn from the tail of a long-tailed distribution \citep{feldmantheory, feldman}, have conflicting properties to their labels or are associated with other labels, such as multiple objects, and most significantly, mislabeled samples. Figure \ref{fig:cherry} shows cherry-picked examples from the 100 highest curvature samples from different vision datasets.  
% Some examples identified by our method are shown in Figure \ref{fig:cherry}.
Consider the images shown for CIFAR100, categories such as `girl' and `baby' and `folding\_chair' and `rocking\_chair' from ImageNet. The images are the same, and they semantically belong to both categories, but would be confusing for single label datasets. Similarly in CIFAR100, images for the classes of `plain', `mountain', and `cloud' have a large overlap. Such similar or overlapping samples can be challenging to catch while creating datasets, and can then get allotted to separate annotators that label them differently. ImageNet labels are further complicated due to multi-object images as is the case with `red\_wine' and `corkscrew' where both the objects are visible in the image, hence creating conflicting but semantically correct labels. Further, when looking at high curvature ImageNet images we also find  other failure modes, such as a non-prototypical image for `sunglasses', mislabelled examples of `cassette'  when clearly it is a `cassette\_player', and conflicting conflicting labels for `chime' and `gong'. 

The quality of datasets impacts not only the accuracy of the network but also encodes hidden assumptions into the datasets, raising issues of bias and unfairness. Underrepresented samples of a class in a dataset can have a significant impact on the performance of the network in real-life settings \citep{bias}. For datasets not vetted by the public, or those that are weakly labeled or have noisy annotations, curvature analysis can help to audit datasets and improve their quality. As an example, while studying curvature properties of CIFAR100 and ImageNet vision datasets we find that nearly half of the top 100 high curvature images correspond to duplicated images \textit{with different labels} in their training sets (Figure \ref{fig:cf100_pairs} and Appendix Figure \ref{fig:imgnet_pairs_appedix}). While the duplication of images in CIFAR100 has been noted before by \cite{recht} and  \cite{purging}, differently labeled duplicated images are, to the best of our knowledge, a novel observation. %When applying the same analysis on ImageNet we found 45 of the top 100 high input loss curvature examples were duplicates but labeled differently \footnote{Further analysis found 14013 ($\sim 1.1\%$ of ImageNet) duplicates in the ImageNet trainset}. \textcolor{red}{I think this footnore raises a lot of questions, whats the point we are making with it?}

In order to validate our measure of curvature quantitatively, we show that our scores achieve a high cosine similarity 
% (\textcolor{red}{can we add some impressive numbers here, either >0.7 with all data or >0.9 with the most memorized data? if not, ignore}
with memorization scores released by \citet{feldman} for CIFAR100 and ImageNet datasets. The baseline scores were calculated from extensive experimentation, and required training thousands of models. In contrast, we can calculate curvature while training only one network, while identifying failure modes that the baseline fails to capture. As a second, independent check, we synthetically mislabel a small proportion ($1-10\%$) of the dataset by introducing uniform noise in the labels and overfit a network to this noisy dataset. We then use our method to calculate the AUROC score for identifying the mislabeled examples. We consistently get high AUROC numbers (between $0.84$ and $1$) for the datasets studied, highlighting that the mislabeled samples were learned with high curvature. 

%However, we emphasize that the purpose of this paper is not to pitch curvature analysis as the best method of finding mislabeled samples in the dataset, but instead to establish it as a good and computationally cheap metric for memorization, and use it in conjunction with other methods to find different failure modes and ensure dataset integrity.

\section{Related Work}
\label{sec:litrev}
Deep neural networks are highly overparametrized, thus suffer from overfitting. Overfitting leads networks to memorize training data to perfect accuracy, while not improving generalization capability.
% as seen by saturated or degrading performance on the test set. 
There has been work showing that 
% unlike classical machine learning algorithms, 
overfitting in neural networks might be benign, and might not harm testing accuracy \citep{benign}. However, overfitting can create other problems such as increased susceptibility to membership inference attacks \citep{membership, canary},
% memorization of data with completely randomized labels \citep{zhang2017understanding} 
and compromised adversarial robustness \citep{adv-overfit}.  
% There have been many methods to counter overfitting, focusing on either regularization of parameters, such as weight decay \citep{wtdecay}, dropout \citep{dropout}, pruning \citep{pruning} or improving training dataset complexity through augmentation techniques\citep{augment} such as cutout \citep{cutout} and mixup \citep{mixup}. 
The extremity of memorization was first noted in neural networks by \cite{zhang2017understanding}. Since then, there has been work in understanding memorization better \citep{closerlook,feldman, feldmantheory, cscores, geom-mem}. \citet{feldman} suggest calculating memorization score by removing a set of examples from the training set and observing the change in their prediction upon training with and without the samples. 
% This also relates to research on the easiness or hardness of examples, or the prototpyical versus outlier samples. 
Related to memorization, research on the easiness or hardness of examples by \citet{rais} and \citet{kf18} utilize the gradient of the sample as the metric of its importance. AUM \citep{aum} considers the difference between the target logit and the next best logit as a measure of importance, and \citet{secondsplit} captures the sample importance in the time it takes to be forgotten when the network is trained with the sample removed.
% from the first split and finetuned in the second split with the example present. 
\citet{carlinidensity} study the prototypicality and memorization of samples based on 5 different correlated metrics. 
% such as confidence score, adversarial vulnerability, ensemble agreement and holdout retraining. 

Methods to counter overfitting include weight decay penalty \citep{wtdecay}, dropout \citep{dropout}  and augmentation \citep{augment}. However, the success of these techniques depends on having a clean and reliable training dataset.
% Dataset cleanliness is an important issue that raises concerns about bias and fairness in real life scenarios.
% There has been some interest in sanity-checking image datasets.
Confident Learning \citep{cleanlab} uses the principles of pruning, ranking and counting with out of prediction probabilities to find mislabeled data. The authors extend it \citep{clean2} to show errors in many common image datasets. \citet{recht} build new test sets for CIFAR and ImageNet to check if there has been unintentional overfitting to the released validation sets. \citet{purging} find many duplicated samples in the CIFAR dataset and create a tool to flag possible duplicates.  We looked at the worst-case examples from these methods and found that they did not outrightly catch duplicated examples with different labels, whereas our method does. 
% We believe that new datasets should be audited with all these methods to catch all possible failure modes.
The curvature of the loss function \textit{with respect to the parameters} is well studied in its connection to the generalization capability of the solution \citep{keskar-large, dinh-sharp, ghorbani-investigation}. 
% A repository of efficient methods to calculate the curvature properties of loss with respect to the weights is made available by \citet{pyhessian}. 
% and can be extended to calculate the loss with respect to data. 
However, far fewer works focus on the properties of curvature of loss \textit{with respect to data}. Prior work in this area has focused on adversarial robustness \citep{fawzi-cure, fawzi-topology}, and on coresets \citep{garg2023samples}. However, in this paper, we propose a new application of input loss curvature as a metric for memorization.
\section{Methodology}
\label{sec:meth}
In this section, we outline how we measure curvature and the computational cost involved. A similar method is often used to calculate the curvature of loss with respect to the network parameters to determine solution stability \citep{dinh-sharp, ghorbani-investigation, keskar-large}. However, we use this method to calculate the curvature of the loss with respect to the input points. It is conceptually similar to the method used for curvature regularization in \citet{fawzi-cure} and  \citet{garg2023samples} with differences in the choice of random variables, normalization, and hyperparameters. Details regarding the hyperparameters are provided in Appendix \ref{supp:hyperparams}.

\subsection{Measuring Curvature}

Let $X \in \mathbb{R}^D$ be the input to a neural network. Let $y \in \{1,2,..,C\}$ be the assigned label for this point, corresponding to the index of the true class among C classes. Let $\hat{y} = f(X,W) \in \mathbb{R}^C$ be the output (pre-softmax logits) of the network with weights W. Let $L(X) = CrossEntropy(\hat{y}, y) \in \mathbb{R}$ be the loss of the network on this data point. We are interested in the Hessian of the loss with respect to $X$, $H(X) \in \mathbb{R}^{d \times d}$, where each element of the matrix is defined by
\vspace{-1mm}
\begin{equation}
    [H(X)]_{i,j} = \left[ \frac{\partial^2 L(X) }{ \partial x_i \partial x_j}  \right] ; i,j = 1,2...,D
\end{equation}
\vspace{-1mm}

Henceforth, we refer to $H(X)$ as $H$ from now, implicitly understanding that it is calculated with respect to datapoint $X$ at a given weight $W$.
The local curvature is determined by the eigenvalues of $H$ \citep{dinh-sharp, ghorbani-investigation, keskar-large}. The sum of the eigenvalues is also the trace of the $H$, and can be calculated using Hutchinson's trace estimator \citep{hutchinson}.

\vspace{-1mm}
\begin{equation}
    Tr(H) = \mathbb{E}_v \left[  v^T H v   \right]
\end{equation}
\vspace{-1mm}

where $v \in \mathbb{R}^D$ belongs to a Rademacher distribution, i.e $v_i = \{+1,-1\}$ with equal probability.
However, we are more interested in themagnitude of the curvature rather than the definiteness, and hence we look at the trace of the square of the hessian, which computes to the sum of the square of the eigenvalues. Since the Hessian is symmetric, we have:
\begin{align}
    Tr(H^2) &= \mathbb{E}_v \left[  v^T H^2 v   \right] \nonumber\\
          % &= \mathbb{E}_v \left[  v^T H^T H v   \right] \nonumber \\
          &= \mathbb{E}_v \left[  (Hv)^T (Hv)   \right] \nonumber \\
          &= \mathbb{E}_v  \lVert Hv \rVert_{2}^{2} \nonumber   \\
          &= \frac{1}{n} \sum_{i=0}^{n} \rVert Hv_i \rVert_{2}^{2}    
\end{align}

where n is the number of Rademacher variables to average over.
Similar to \citet{fawzi-cure} and \citet{garg2023samples}, we use finite step approximation to calculate this efficiently.
\begin{align}
    Hv &= \frac {1}{h} \left[ \frac{\partial L(x + hv)}{\partial x} - \frac{\partial L(x)}{\partial x} \right] \\
    Hv &\propto \frac{\partial  \left(L(x+hv) - L(x)\right)}{\partial x} 
\end{align}

We drop constants as we are only interested in the relative curvatures of datapoints. For our final curvature estimate, we average curvature over all training epochs, T, to give reliable results. Putting this together, we have the curvature estimator of a datapoint $X$ at any epoch as:
\begin{align} \label{eq:est}
    Curv(X) = \frac{1}{nT}  \sum_{t=1}^{T} \sum_{i=0}^{n} \left \lVert \frac{\partial (L(x+hv) - L(x)) }{\partial x}  \right \rVert_{2}^{2}
\end{align}

\subsection{Computational Cost}
The cost of performing the calculation in Equation \ref{eq:est} is $O(n)$ forward and backward passes, where $n$ is the number of Randemacher variables $v$ to average.  The backward passes can be parallelized as they are performed at a static $W$. In case of limited compute capacity, curvature could be estimated every few epochs. Note that we do not explicitly compute the Hessian or any second-order gradient despite relying on a second-order metric.

\section{Experiments and Discussion}
\label{sec:qual_res}

% Visualize low and hig curv samples
% Explore properties of high curvature samples, how they develop during training, their connection with overfitting and with FZ scores.

In this section, we present qualitative and quantitative results on MNIST \citep{mnist}, FashionMNIST \citep{fmnist}, CIFAR10/100 \citep{cifar}, and ImageNet \citep{ILSVRC15} datasets to support our claim that curvature can be used to measure memorization. 
The histograms of curvatures are shown in Figure \ref{fig:hist}, and we can see that they reflect the long-tailed nature of natural images.
% first visualize what the typical low-curvature and high-curvature samples look like for MNIST, FashionMNIST, and the CIFAR datasets. We then show how values of curvature change during training. We also explore the role of weight decay/regularization on curvatures during training. \citet{feldman} measure memorization of samples independently of training statistics, by removing an example from the dataset and measuring the change in probability of its prediction. Those scores are unlikely to have a non-meaningful or spurious correlation to curvature, and hence we use them to support our claim that curvature can be used to measure memorization. In particular, we show that the average curvature of samples identified as being memorized by \citet{feldman} is higher than the average curvature of the entire trainset, at all epochs during training. Further, we measure the epoch-wise cosine similarity between sample curvatures and the scores released by \citet{feldman} for CIFAR100 at different epochs. We note that the match is very good in the epochs right after overfitting, but then degrades in the later epochs as overfitting progresses. However, the cumulative curvature remains a good match ($>0.8$ cosine similarity). The curvature analysis leads to a novel discovery about the CIFAR100 dataset. 36 out of the top 60 images with the highest curvatures are duplicate images with different labels. Next, we quantify how well our method captures synthetically mislabeled examples. Similarly, 45 of the top 100 examples in ImageNet. 
We use ResNet18 \citep{resnet} for all our experiments. %all datasets, appropriately modified. 
Curvature for all datasets except ImageNet is calculated every epoch and averaged. For ImageNet we average curvature calculated every 4 epochs for computational ease.
% We encourage practitioners with extensive GPU compute availability to calculate scores per epoch for more reliable results.
Details of the setup are given in Appendix \ref{supp:arch}. 
%Isha: only show 1 pic for histogram
This section first shows qualitative results by visualizing high curvature samples, and showing that they are long-tailed or mislabeled images in section \ref{4.1}. We then validate our score quantitatively in section \ref{4.2} by comparing against a baseline and in section \ref{4.3} by synthetically mislabeling data and showing that it gets learned with high curvature. Lastly, in section \ref{4.4}, we study how our measure of curvature evolves during training for further insight, and so that practitioners with limited compute can better choose limited epochs for curvature calculation.

\begin{minipage}[b!]{\textwidth}
  \begin{minipage}[c]{0.46\textwidth}
    \centering
    \includegraphics[width=0.85\textwidth]{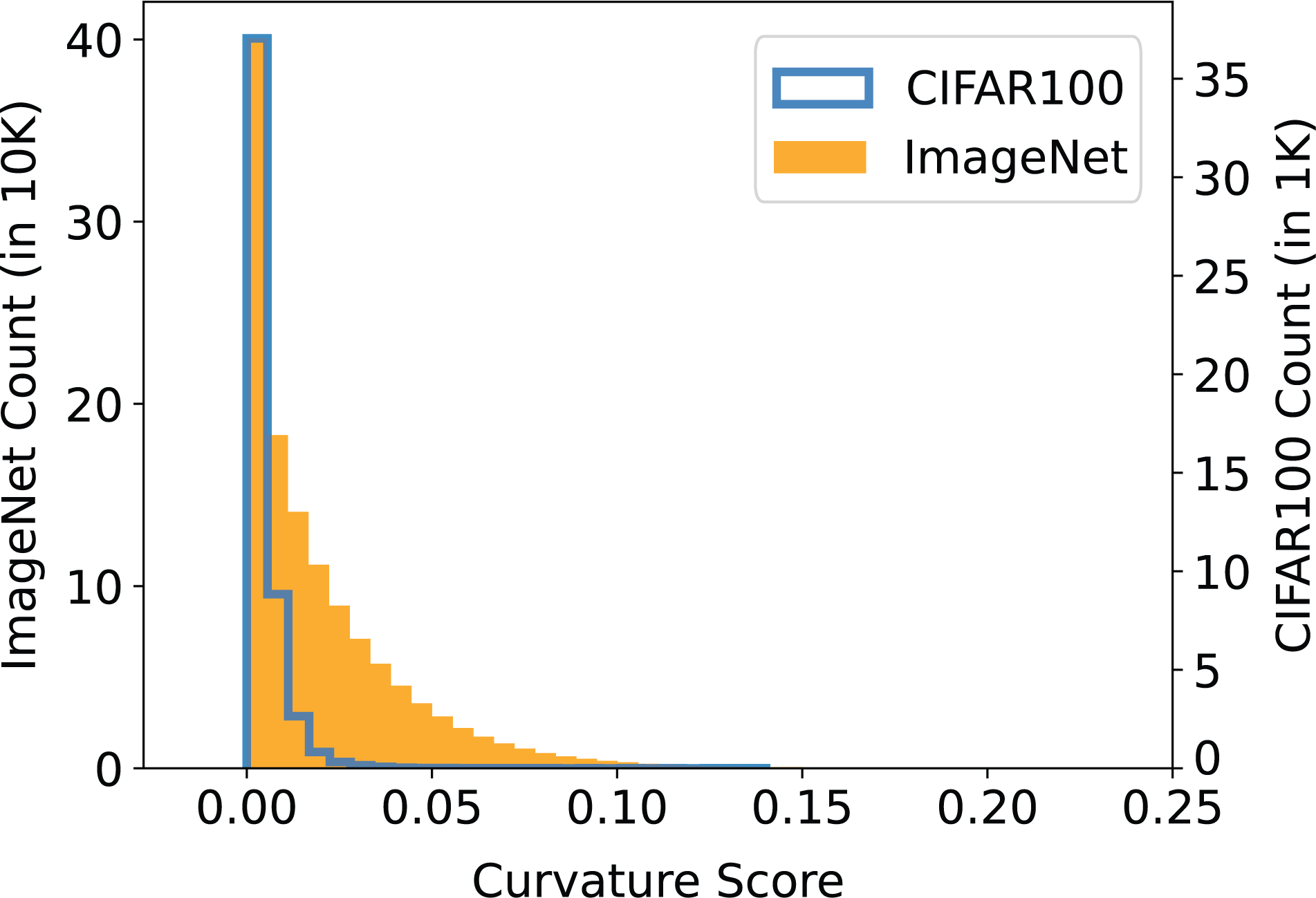}
    \captionof{figure}{Histogram of curvature scores}
    \label{fig:hist}
  \end{minipage}
  \hfill
  \begin{minipage}[c]{0.53\textwidth}
    \centering
    \captionof{table}{Cosine Similarity (CS) between curvature and FZ scores with and without weight decay (WD). Top-K CS is the CS of the top 5,000 (and 50,000) FZ score samples of CIFAR100 (and ImageNet).}
    \label{tab:wd_effect}
    \renewcommand*{\arraystretch}{1.2}{
    \begin{tabular}[b]{|c|c|c|c|} \hline
    Dataset                   & WD & CS & Top-K CS\\ \hline
    \multirow{2}{*}{CIFAR100} & 1e-04            & 0.82              &       0.90             \\
                              & 0.0     & 0.73             & 0.82                   \\ \hline
    \multirow{2}{*}{ImageNet} & 1e-04            & 0.72              & 0.87               \\
                              & 0.0     & 0.66              & 0.82    \\ \hline          
    \end{tabular}}
  \end{minipage}
\end{minipage}

% \begin{minipage}[b!]{\textwidth}
%   \begin{minipage}[c]{0.49\textwidth}
%     \centering
%     \includegraphics[width=0.8\textwidth]{icml2023/pics/mnist_wd0_change60_hist.png}
%     \captionof{figure}{Histogram of MNIST trainset curvature. Curvature of 60 corrupted samples marked.}
%     \label{fig:mnist_hist}
%   \end{minipage}
%   \hfill
%   \begin{minipage}[c]{0.5\textwidth}
%     \centering
%     \captionof{table}{Cosine Similarity (CS) between curvature and FZ scores with and without weight decay (WD). Top-K CS is the CS of the top 5,000 (and 50,000) FZ score samples of CIFAR100 (and ImageNet).}
%     \label{tab:wd_effect}
%     \renewcommand*{\arraystretch}{1.2}{
%     \begin{tabular}[b]{|c|c|c|c|} \hline
%     Dataset                   & WD & CS & Top-K CS\\ \hline
%     \multirow{2}{*}{CIFAR100} & 1e-04            & 0.82              &       0.90             \\
%                               & 0.0     & 0.73             & 0.82                   \\ \hline
%     \multirow{2}{*}{ImageNet} & 1e-04            & 0.72              & 0.87               \\
%                               & 0.0     & 0.66              & 0.82    \\ \hline          
%     \end{tabular}}
%   \end{minipage}
% \end{minipage}

\vspace{-1mm}
\subsection{Visualizing Low and High Curvature Samples}
\label{4.1}

\begin{figure}[t]
    \centering
    \subfloat{
        %\label{fig:mnist_both}
        \centering
        \includegraphics[width=0.47\columnwidth]{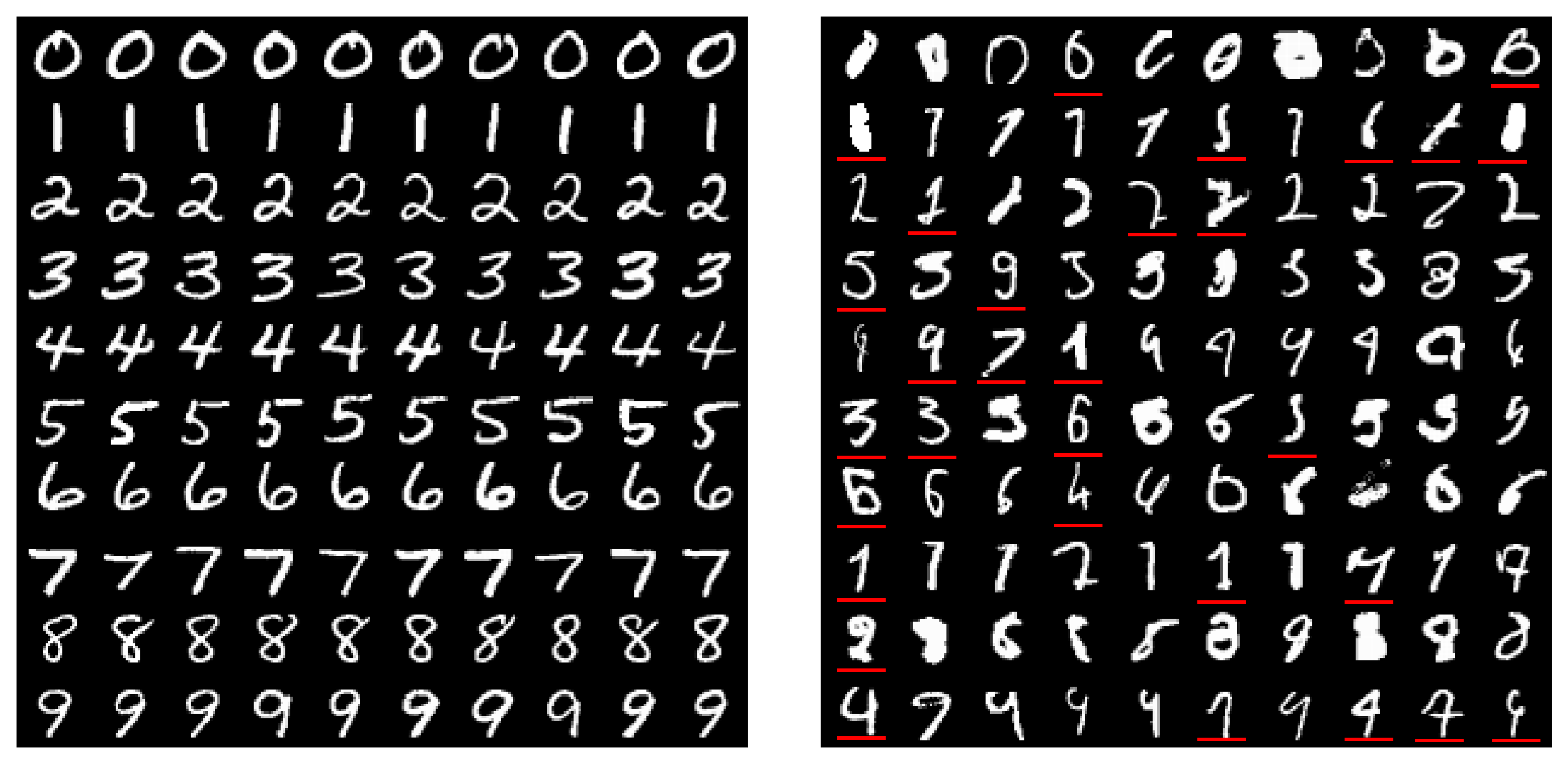}
    }
\hfill
    \centering
    \subfloat
    {   %\label{fig:fmnist_both}
    \includegraphics[width=0.48\columnwidth]{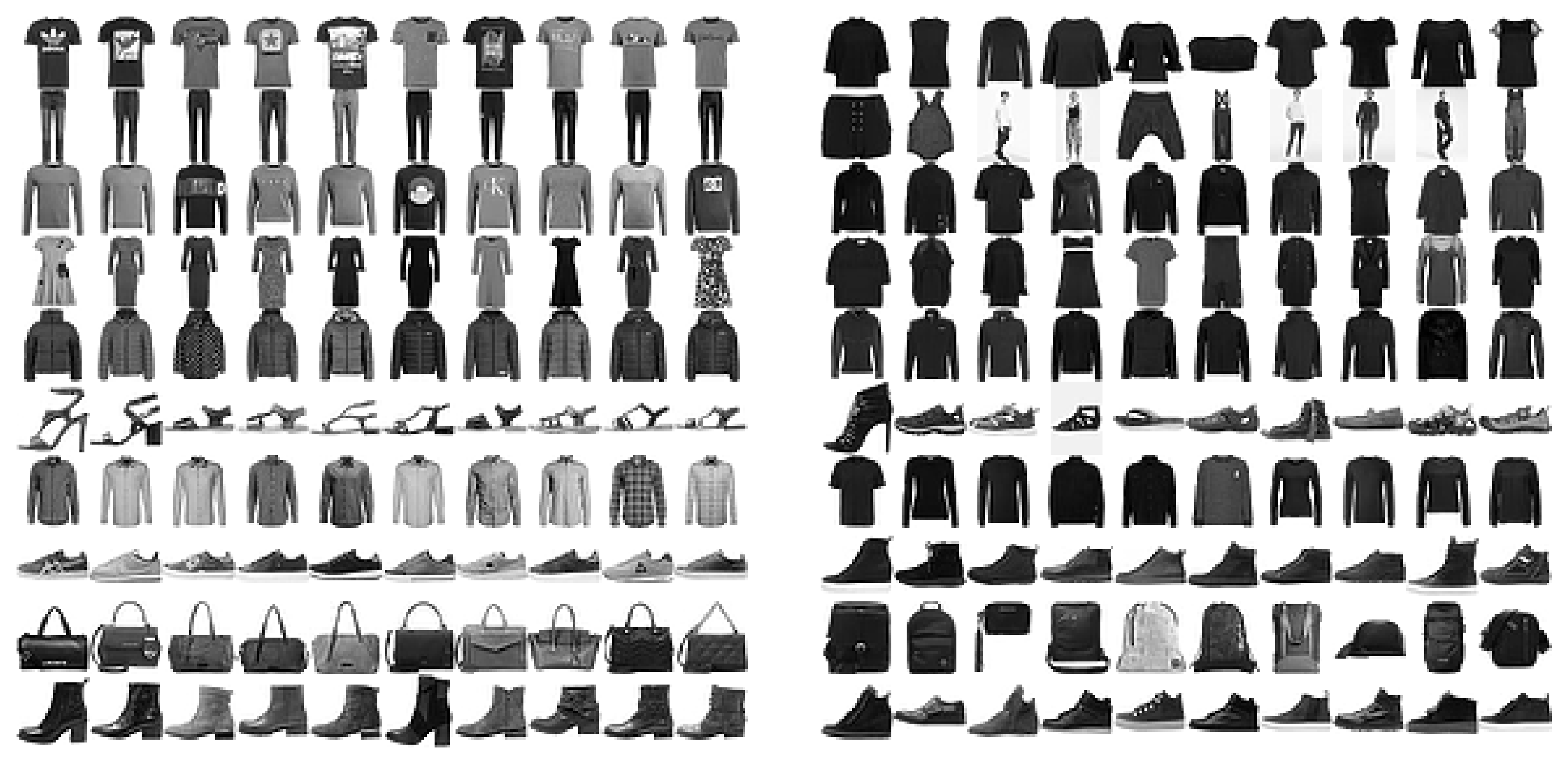}
    }
\caption{Low (left) and high (right) curvature samples of MNIST and FashionMNIST training data}
\label{fig:mnist_fmnist_hiker_viz}
\vspace{-2mm}
\end{figure}

In Figure \ref{fig:mnist_fmnist_hiker_viz}, we visualize the 10 lowest and highest curvature training samples for each class of MNIST and FMNIST respectively. These are examples sorted by curvature scores averaged over all training epochs of a modified ResNet18 architecture, trained without weight decay and allowed to be fully memorized (100\% training accuracy). 
We see that low curvature samples appear to be easily recognizable as belonging to their class. 
% This was exploited in \citet{garg2023samples} to make performant small-sized coresets. 
Whereas, high curvature samples, are made of both long-tailed (harder samples or rare occurrences) and mislabeled samples. The mislabeled samples in MNIST are underlined in red, and we note that the remaining high curvature samples are also  ambiguous. In particular, we see high overlap between classes `4' and `9', and `1' and `7'.

% \begin{figure}[t]
%     \centering
%     \captionsetup[subfigure]{labelformat=empty}
%     \subfigure{
%         %\label{fig:mnist_both}
%         \centering
%         \includegraphics[width=0.47\columnwidth]{icml2023/pics/mnist_nocorr.png}
%     }
% \hfill
%     \centering
%     \subfigure
%     {   %\label{fig:fmnist_both}
%     \includegraphics[width=0.48\columnwidth]{icml2023/pics/fmnist_nocorr.png}
%     }
% \vspace{-2mm}
% \caption{Low (left) and high (right) curvature samples of MNIST and FashionMNIST training data}
% \label{fig:mnist_fmnist_hiker_viz}
% \vspace{-2mm}
% \end{figure}

The classes for FashionMNIST are 0:T-shirt/top, 1:Trouser, 2:Pullover, 3:Dress, 4:Coat, 5:Sandal, 6:Shirt, 7:Sneaker, 8:Bag, 9:Ankle boot. We see that examples seem to have significant overlap for classes `T-shirt', `Pullover' and `Coat', and among `Ankle boots' and `Sneakers'. We also see long-tailed trends in the row of `handbags', with some uncommon handbags being represented in the high curvature case. Conflicting samples are apparent in the row for pants, as there are full body pictures of models, wearing both pants and shirts. This analysis reveals a bias in the FashionMNIST dataset: darker clothes with lower contrast tend to show up as high curvature samples, whereas in the real world, this may not be the case. 
% Biases in performance based on image contrast, brightness, or color profiles in the real-world raise questions about fairness. 
% Performing differently based on color is an issue that can create significant problems with fairness of outcomes \textcolor{red}{maybe rephrase, maybe find a ref?}
The corresponding pictures for the training sets of CIFAR10, CIFAR100 and ImageNet are shown in Appendix \ref{supp:train_viz}. We check the validation data as well by training and overfitting on the validation sets, and the results are shown in Appendix \ref{supp:val_viz}.

\subsection{Comparing Against a Baseline: FZ scores}
\label{4.2}

%removed mu because we are not talking about epcohwise result here
Here, we are interested in an independent measure of memorization that does not utilize training dynamics. The most suitable metric for comparison comes from \citet{feldman}, who remove datapoints from the dataset one at a time, train a new network on the altered dataset, and measure the change in prediction on the datapoint as the sample memorization score. These scores are likely to be independent of spurious correlations to curvature that other methods such as confidence of prediction might have, and hence serve as a good baseline. The authors released scores for the CIFAR100 and ImageNet dataset on their site\footnote{https://pluskid.github.io/influence-memorization/}, and we refer to these as FZ scores. We calculate the cosine similarity of our curvature with the FZ scores in Table \ref{tab:wd_effect}. \textbf{We achieve a cosine similarity of 0.82 and 0.72 for CIFAR100 and ImageNet respectively when we use weight decay, and 0.90 and 0.87 on the most memorized samples} (see Table \ref{tab:wd_effect}). These scores are a high match since the vector we are using is of dimension 50,000 in the case of CIFAR100, and $1,268,356$ in the case of ImageNet (number of training samples in the dataset). We emphasize that we achieve this match with only 1 training run, whereas FZ scores required training thousands of models.

\begin{figure}[bt!]
\centering
\includegraphics[scale=0.35]{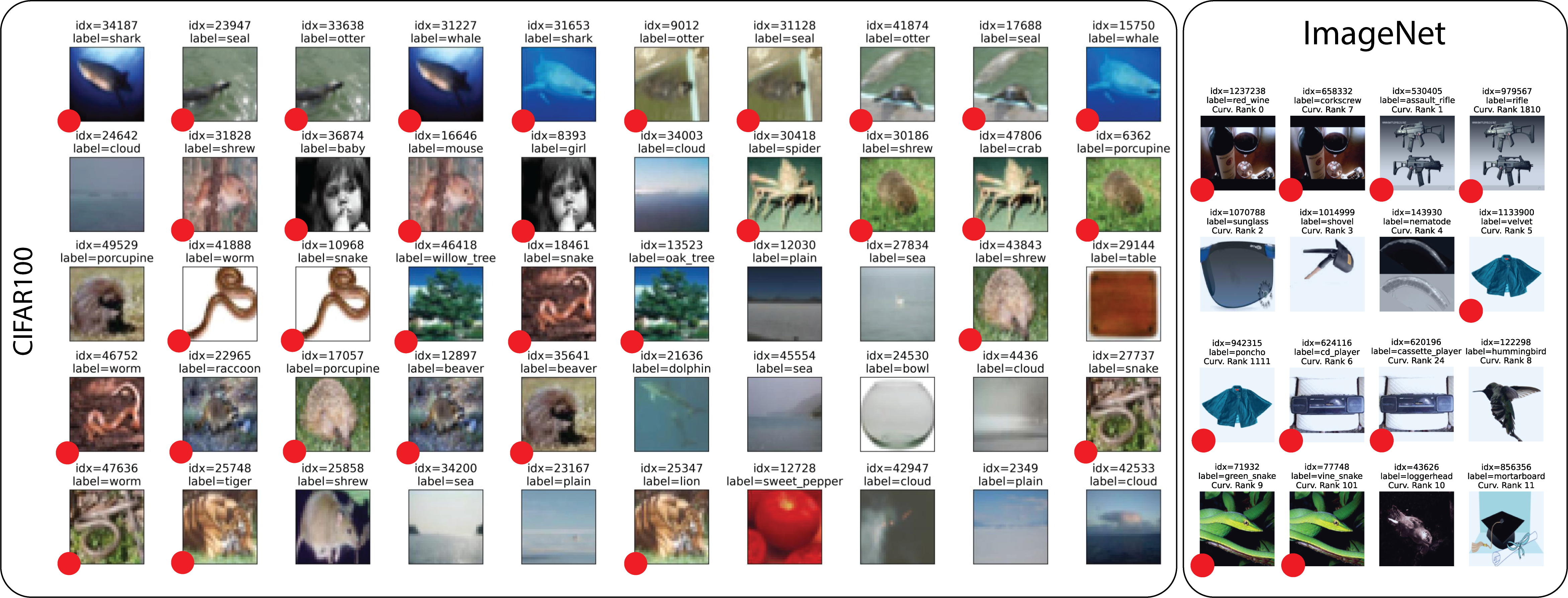}
\caption{50 highest curvature samples from the training set of CIFAR100 and ImageNet, index, label (and curvature rank for ImageNet) shown above each image. We highlight duplicated samples with differing labels with a red dot. Extended versions in Appendix, Figures \ref{fig:cf100_pairs_appendix} and \ref{fig:imgnet_pairs_appedix}.}
\label{fig:cf100_pairs}
\vspace{-4mm}
\end{figure}

Counter intuitively, we note that the cosine similarity drops by $\sim0.1$ when not using weight decay. To understand why, we visualize the examples with the highest curvature without weight decay in Figure \ref{fig:cf100_pairs} for CIFAR 100 and ImageNet. We note that \textbf{36 out of the 60 highest curvature samples for CIFAR100, and 45 out of top 100 for ImageNet are duplicated pairs with conflicting labels.} These are marked with a red dot. In contrast, FZ scores do not catch duplicated samples. Similarly, weight-regularized, highest curvature samples only catch a few of the duplicated samples, as regularization would deter the network from memorizing these samples. The 100 most memorized samples from FZ scores and the highest curvature samples for training with weight decay are shown in Appendix \ref{supp:cf100_worst}. These duplicated samples are indeed memorized during training but possibly missed by FZ scores due to the fact that they do not train until complete memorization due to the computational expense of training thousands of models. Additionally, FZ scores are calculated by removing a proportion of the dataset at a time and training on the remaining samples. Duplicates can compromise the reliability of these scores. We note that there are many other duplicate samples that our method does not identify \citep{purging}, since these duplicate samples have the same label and do not pose a boundary conflict. They are also not likely to be memorized.  

To summarize, one of the reasons we do not get near perfect match with FZ scores (and why we get a higher match with weight decay) is because \textbf{we catch a novel failure mode that the FZ scores fail to, despite being $\sim3\times$ more computationally expensive.}
% The better match with FZ scores with regularization hints that we might be able to capture corrupted labels better with regularization. 
For the sake of completeness, we recommend that practitioners try both settings (with and without weight decay) to try and catch the different kinds of boundary conflicts that may be revealed via curvature analysis. We do a more detailed analysis in section \ref{4.4}, where we show the cosine matches per epoch.

\vspace{-2mm}
\subsection{Synthetic Label Corruption}
\label{4.3}
To provide further evidence of curvature as a memorization metric, we devise an experiment to measure how well our method captures synthetically mislabeled examples, since they are most likely to be memorized. We randomly introduce noise into a proportion of the labels, uniformly changing its class label to a different class label. The same proportion of labels in each class are corrupted. We then train on these samples and measure the curvature scores. 
First,  we randomly corrupt the label of 60 images per class of MNIST (corrupting 1\% of the dataset), and train until full memorization, i.e. 100\% training accuracy, indicating that the mislabeled samples were memorized. We plot the histogram of the cumulative curvature scores and mark the scores of the corrupted ones with a black $\times$ in Figure \ref{fig:mnist_hist}. The training dataset consists of 60,000 samples and we can see that most of the samples have very low curvature. However, we see that \textbf{the curvature scores of the corrupted examples are among the highest,} confirming the link between curvature and memorization.

% \begin{figure}[!htbp]
% \centering
% \subfloat[Loss curves and average trainset curvature estimates on CIFAR100.]{\label{fig:overfitting}
% \centering
% \includegraphics[width=0.46\linewidth]{icml2023/pics/overfit.png}
% }
% \hfill
% \centering
% \subfloat[Histogram of curvature scores for ImageNet and CIFAR100 datasets.]{
%   \centering
%   \includegraphics[width=0.48\linewidth]{icml2023/pics/histogram.png}
%   \label{fig:histogram}
% }
% \caption{(a) Curvature, validation and training loss development. (b) Histogram of curvature score showing long-tailed nature obtained on ResNet18 trained on ImageNet and CIFAR100.}
% \vspace{-2mm}
% \end{figure}

To do a more exhaustive study, we sort samples by their curvature estimate and report the AUROC (Area under the ROC curve) scores for separating the corrupted samples from the clean ones in Table \ref{tab:auroc}, for corruptions ranging between 1-10\% of 3 widely used vision datasets. The best results are highlighted in bold red. Since the uncorrupted class has far higher number of samples than the corrupted class for this binary classification case, ROC curves are more reliable than the precision-recall curves. As shown later while studying curvature dynamics in section \ref{4.4}, the scores averaged over training are stable when using weight decay, and hence we use a weight decay of $10^{-4}$ for all models for these experiments.
\begin{table}[!t]

\renewcommand*{\arraystretch}{1.0}
\centering
\caption{AUROC for identifying corrupted samples with synthetic label noise for different corruption percentages of datasets. The best results are shown in red. Inconf. refers to inconfidence score, and CL refers to Confident Learning \citep{cleanlab}.}
\vspace{-2mm}
\begin{tabular}{|c|ccc|ccc|ccc|}
\hline
             & \multicolumn{3}{c|}{MNIST}      & \multicolumn{3}{c|}{CIFAR10}                                            & \multicolumn{3}{c|}{CIFAR100}                                           \\ \hline
            Corr. & Inconf. & CL                          & Curv.                     & Inconf. & CL                            & Curv.                     & Inconf. & CL                            & Curv.                     \\ \hline
1\%           & 99.4\%  & 99.7\% & \textbf{\color[HTML]{FF0000} 100.0\%}  & 84.2\%  & 92.4\%                        & \textbf{\color[HTML]{FF0000} 97.4\%} & 85.0\%  & 83.1\%                        & \textbf{\color[HTML]{FF0000} 90.9\%} \\
2\%            & 99.0\%  & 99.3\% & \textbf{\color[HTML]{FF0000} 100.0\%} & 82.5\%  & 93.1\%                        & \textbf{\color[HTML]{FF0000} 96.6\%} & 84.0\%  & 84.2\%                        & \textbf{\color[HTML]{FF0000} 89.6\%} \\
4\%            & 98.4\%  & 99.1\% & \textbf{\color[HTML]{FF0000} 99.9\%}  & 81.8\%  & 93.4\%                        & \textbf{\color[HTML]{FF0000} 95.5\%} & 83.5\%  & 85.3\%                        & \textbf{\color[HTML]{FF0000} 88.3\%} \\
6\%            & 97.7\%  & 98.9\% & \textbf{\color[HTML]{FF0000} 99.9\%}  & 81.6\%  & 93.4\%                        & \textbf{\color[HTML]{FF0000} 94.4\%} & 83.6\%  & 86.3\%                        & \textbf{\color[HTML]{FF0000} 86.8\%} \\
8\%            & 97.1\%  & 98.9\% & \textbf{\color[HTML]{FF0000} 99.9\%}  & 81.5\%  & 93.5\%                        & \textbf{\color[HTML]{FF0000} 94.1\%} & 83.6\%  & 84.3\%                        & \textbf{\color[HTML]{FF0000} 85.7\%} \\
10\%           & 96.1\%  & 98.9\% & \textbf{\color[HTML]{FF0000} 99.9\%}  & 81.5\%  & \textbf{\color[HTML]{FF0000} 93.4\%} & 92.9\%                        & 83.4\%  & \textbf{\color[HTML]{FF0000} 84.6\%} & 84.3\%                        \\ \hline
\end{tabular}
\label{tab:auroc}

\end{table}

% with curvature averaged over all epochs. 
% Scores for curvature in this section are averaged over all training epochs.
To compare against more baselines, we also provide AUROC results using inconfidence \citep{carlinidensity} and Confident Learning \citep{cleanlab}.
Inconfidence ($=1-$ confidence) was explored as one of the 5 metrics by \citet{carlinidensity} to identify long tail and outlier examples. 
% Another measure of conflict at boundary can be the confidence of the sample, with most samples classified very confidently, and the ones with conflicting boundaries would be classified with lesser confidence. This was also explored as one of the 5 metrics in \citet{carlinidensity}. 
% We compute AUROC scores for sorting by inconfidence values as well. We also compare with a non-boundary related metric, that of Confident Learning (CL) \citep{cleanlab}. 
Confident Learning Confident Learning (CL) \citep{cleanlab} involves 4 sets of cross-validation experiments 
% to get out of prediction probabilities. CL
and identifies examples as error-prone after a certain threshold based on counting and ranking principles.
% so we give a score of zero to all the samples it does not identify. 
%The samples it does identify are ranked and the corresponding AUROC scores are reported.
% , and we keep that rank intact for the purpose of measuring AUROC scores. 
The experiments for CL were performed using the code provided by the authors. From Table \ref{tab:auroc}, we conclude that our method is more informative than just confidence values on all datasets and outperforms cleanlab on most corruptions and datasets. \textbf{We note that the AUROC score of our method on all datasets except CIFAR100 is above 92.9\%, and above 84.3\% on CIFAR100.} We wish to emphasize here that we do not claim that curvature is the best way of finding mislabeled examples. Our primary motivation is to show that curvature is a good measure of sample memorization, and exploiting it can find samples with failure modes that other methods might miss. We recommend that curvature analysis should be used in conjunction with other checks, for a holistic view of dataset integrity.

\subsection{Curvature Dynamics During Training}
\label{4.4}
\begin{figure}[!t]

\subfloat[Average trainset, testset and memorized trainset curvatures on CIFAR100.]{\label{fig:memorization}
\centering
\includegraphics[width=0.46\linewidth]{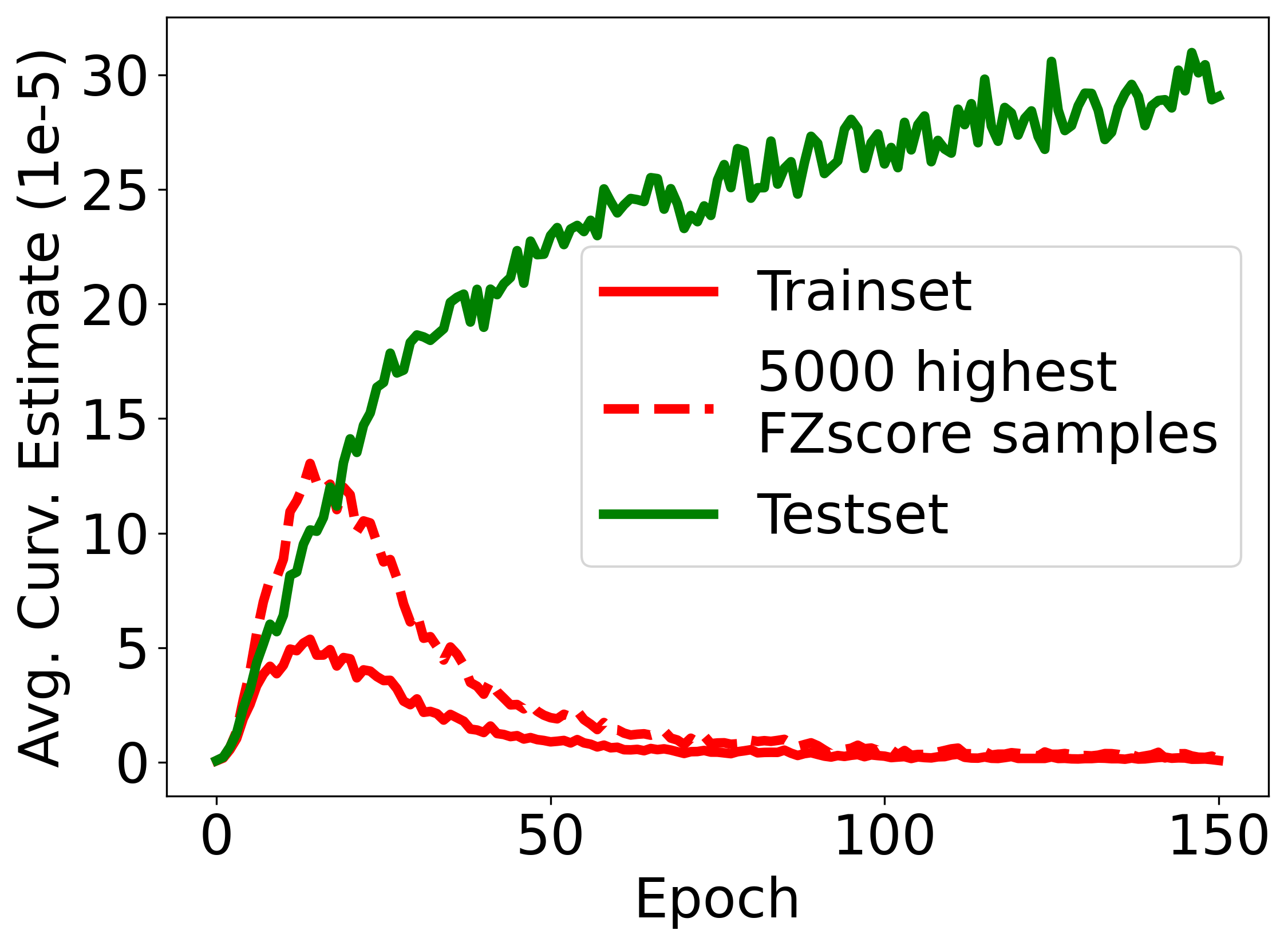}
}
% \caption{Curvature dynamics, ResNet18, CIFAR100}
% \label{fig:overfit_memo}
\hfill
\centering
\subfloat[Average trainset, testset and memorized trainset curvatures on ImageNet.]{\label{fig:memorization_img_net}
\centering
\includegraphics[width=0.48\linewidth]{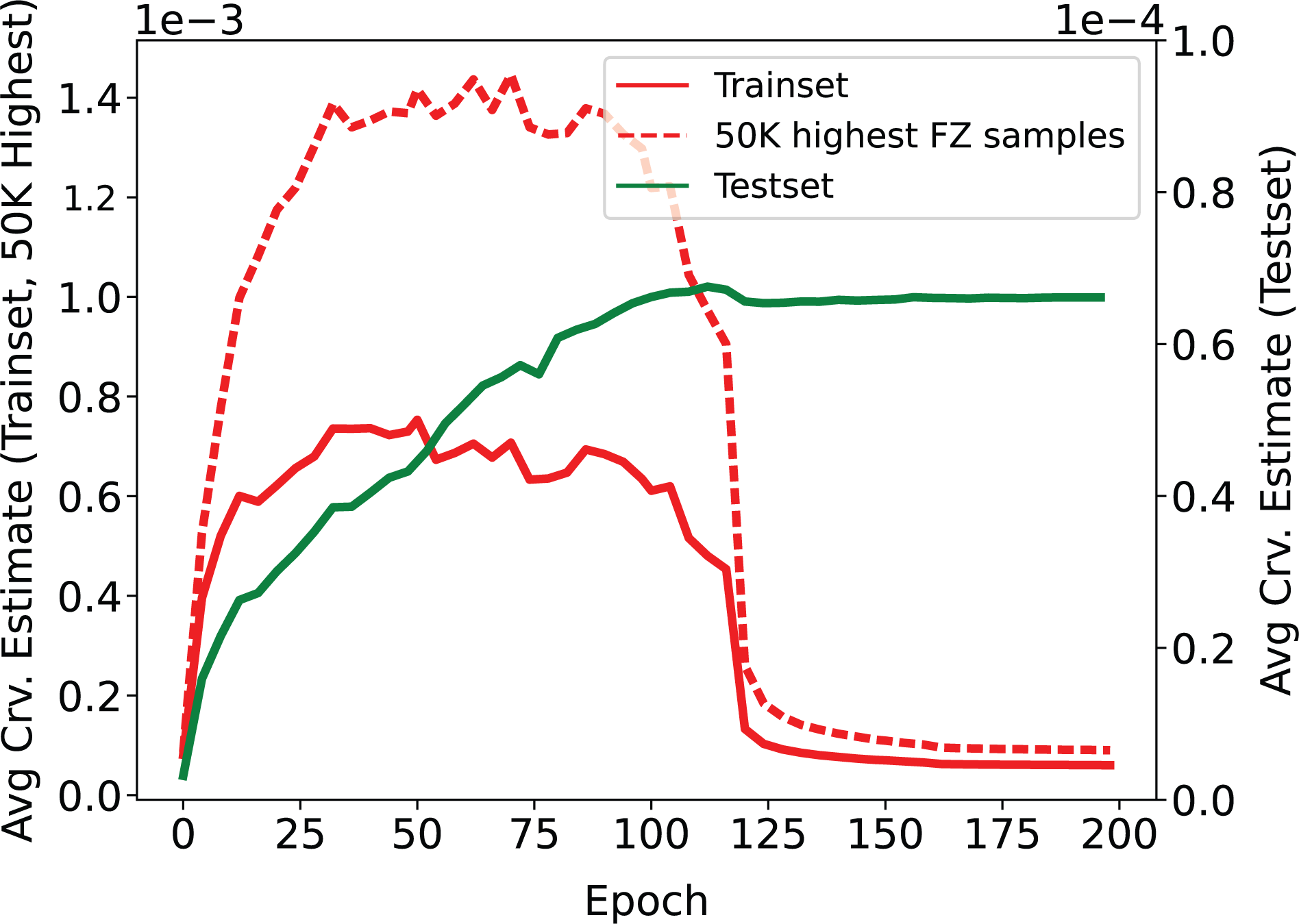}
}
\caption{Curvature dynamics for CIFAR100 and ImageNet. Note for ImageNet we evaluate every 4 epochs for compute efficiency.}
\vspace{-2mm}
\end{figure}

Until now, we have studied curvature score as the average curvature of each training epoch. In this section, we study how the curvature of the training set develops as training progresses, to motivate why averaging curvature scores over training is needed rather than calculating it at the end of training. We study \textit{per-epoch} curvature now, different from the definition of curvature score in Equation \ref{eq:est}. In Figure \ref{fig:overfitting}, we plot the training and validation losses and the per-epoch curvature score averaged for the training set of CIFAR100 on ResNet18. We can see that the network starts to overfit very early, around epoch 20. We see an odd trend in the per-epoch curvature: The phase before overfitting is marked by large  curvature growth until overfitting takes over and the curvature starts to decrease. The same trend is also observed on ImageNet (see Figure \ref{fig:memorization_img_net}).

To understand why this happens, we created a toy example in 2D for visualization. Training and setup details can be found in Appendix \ref{supp:2d}. The top left of Figure \ref{fig:logit_space} shows the training and test data, which consist of 2 classes marked in red and blue sampled from a spiral dataset\footnote{Inspired by the TensorFlow playground     https://playground.tensorflow.org}. The testing data is abundant (100 points for each of the 2 classes) and noise-free, representing the true  distribution. However, the training data consists of only 15 points for each of the 2 classes. 

\begin{minipage}[b!]{\textwidth}
  \begin{minipage}[c]{0.49\textwidth}
    \centering
    \includegraphics[width=0.8\textwidth]{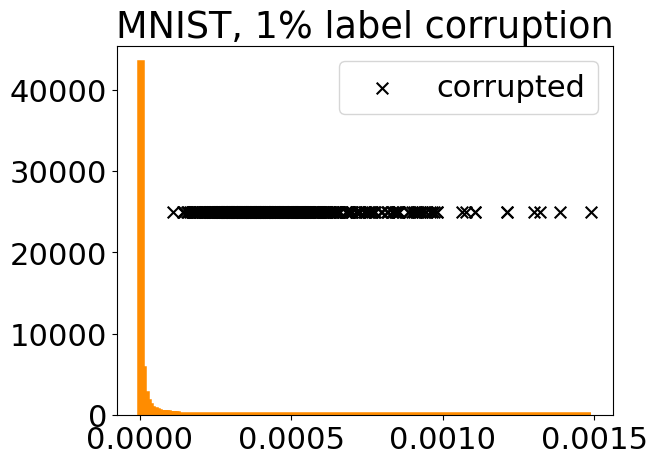}
    \captionof{figure}{Histogram of MNIST trainset curvature. Curvature of 60 corrupted samples marked}
    \label{fig:mnist_hist}
  \end{minipage}
  \hfill
  \begin{minipage}[c]{0.48\textwidth}
    \centering
    \includegraphics[width=0.8\textwidth]{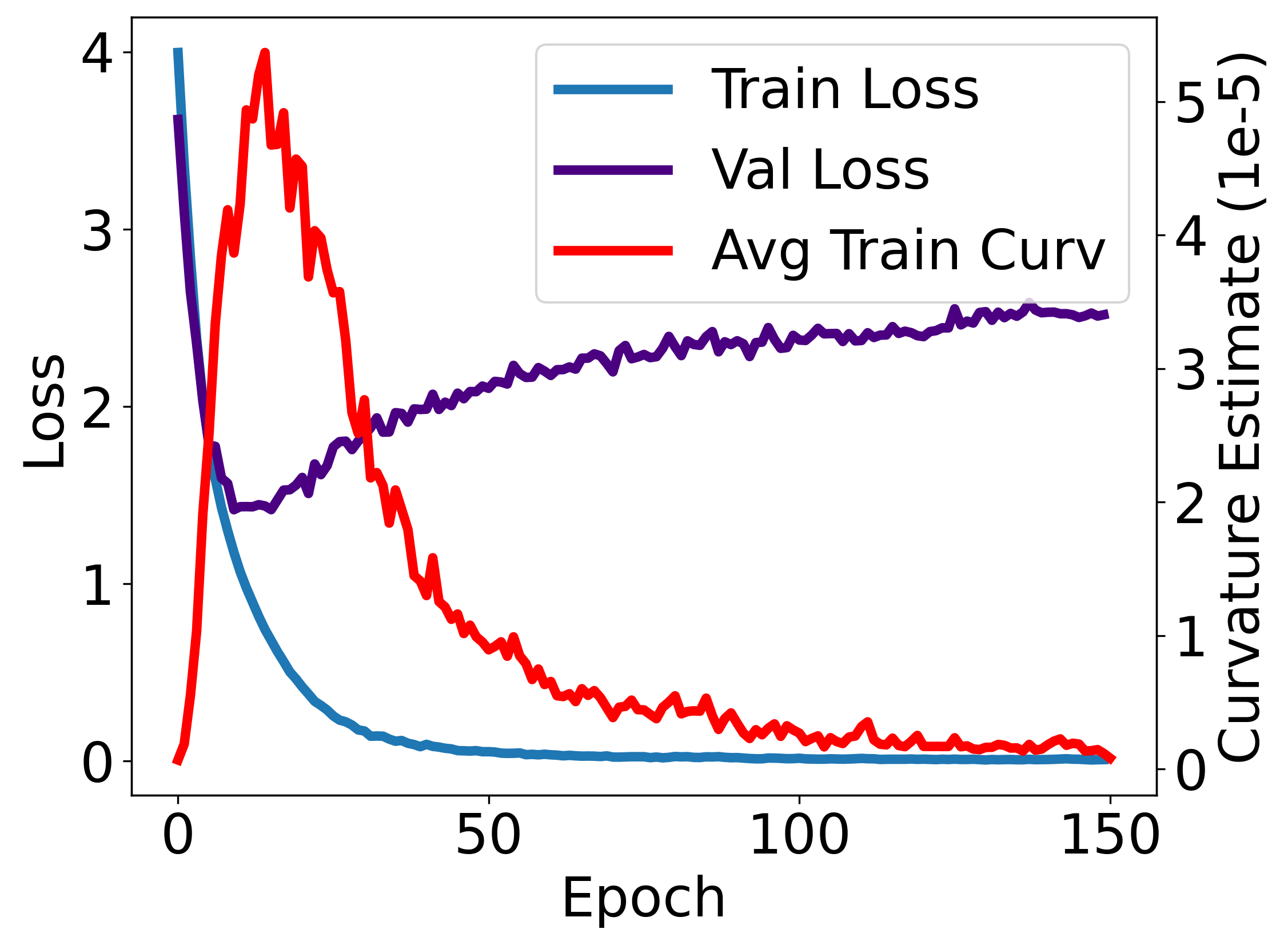}
    \captionof{figure}{Loss curves and average trainset curvature estimates on CIFAR100}
    \label{fig:overfitting}
  \end{minipage}
\end{minipage}

Further, 30\% of the training data has noise added to it to encourage the network to memorize these samples. The top right of Figure \ref{fig:logit_space} shows the loss curves for training and testing data (cross-entropy loss). At the bottom, we show snapshots of the output space (logits) at the top and the input space at the bottom. For the output space, we show the $x$ and $y$ axes, corresponding to each class. The decision boundary in the output space is the line $y=x$, shown running diagonally. The input space at the bottom visualizes the probability heatmap of each point in the grid belonging to the red class.
% completely generalizing. 

\begin{figure}
  \centering
  \includegraphics[width=0.8\linewidth]{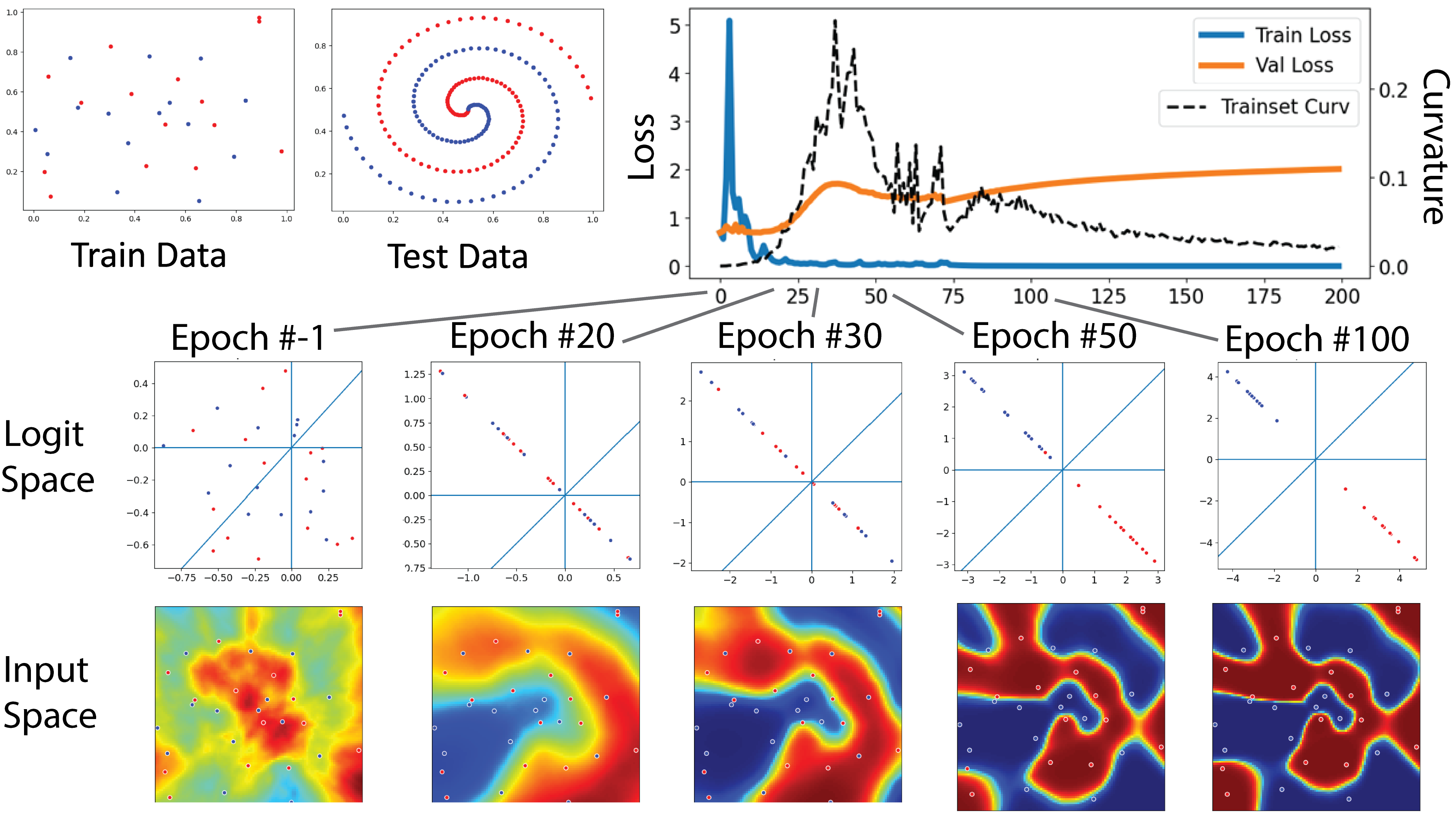}
  \caption{Visualization of curvature of decision boundary around data points. Train and test datasets are shown on the top left, and loss curves on the top right. The bottom figures show the logit space and the decision boundary with input points in the input space at different epochs of training.}
  \label{fig:logit_space}
\vspace{-4mm}
\end{figure}

From the final learned boundaries in the input space, we note that the regions without conflicting data points have largely linear decision boundaries, with complicated boundaries manifesting where data points of both classes are closely intertwined. From the loss and curvature plots on the top right, we can see the same increase and decrease of curvature as before. 
% This becomes clearer as we understand the dynamics of the logits and the softmax layer acting upon them in the cross entropy loss. Early epochs show large changes in the curvature around points in the input space. 
To understand why, we look at the input and the logit space. From the logit space visualization, we see that the focus of early epochs is to get all data points on the correct side of the boundary. Up to this point, the curvature increases. At this point, even though all the datapoints are on the correct side, the cross entropy loss can never go to zero, and is further minimized by maximally separating the correct class logits from the others. This margin maximization is evident in the last three snapshots: the logits get further and further apart, while little change is seen in the input space. This is where the curvature measure starts to drop. As the logits get further apart from the boundary, the softmax layer included in the cross entropy loss makes very confident predictions. Thus, small perturbations have very little impact as the softmax output is already saturated, and hence we see little change in the gradients, required to measure curvature. 

 We emphasize that this trend does not affect the results discussed in sections \ref{4.1}. \ref{4.2} and \ref{4.3}, as we do not compute our curvature score per-epoch, but as the average. Averaging the scores across epochs allows us to distill all the epoch-specific variations into one score and allows us to be oblivious to the exact statistics of learning, including when overfitting began and when the curvature scores started to change trend. However, understanding this trend may help practitioners gain insight into important epochs to average over, in the case of limited compute. Additionally, we are interested only in the relative ranking of the samples. As can be seen from Figure \ref{fig:memorization} and Figure \ref{fig:memorization_img_net}, the curvature of the 5,000 (50,000) most memorized samples as per FZ scores always remains higher, in each epoch for CIFAR100 (ImageNet)
 
Another interesting observation is that the curvature of validation samples keeps increasing throughout training. From Figures \ref{fig:memorization} and \ref{fig:memorization_img_net}, we can see that the prolonged overfitting that causes the training set curvature to decrease due to margin maximization caused by cross entropy loss does not affect the validation samples. This is expected, since we do not minimize loss on the validation or test samples. This affords us some insight into two unrelated mechanisms. First, the sensitivity of test samples to perturbations keeps increasing with training epochs, and this supports the fact that early stopping helps lower adversarial vulnerability \citep{adv-overfit}. Second, longer training make the model more susceptible to membership inference attacks, wherein attackers try to identify the samples used to train the network \citep{membership}. In a white box scenario, measuring the curvature of a sample would help an attacker identify if the sample was used to train the network.

To round up these results, we show the cosine similarity match between per-epoch curvature scores and FZ scores, along with the cosine similarity between the cumulative curvature scores and FZ scores in the Appendix (Section \ref{sec:epoch_ise_appndx} and Figures \ref{fig:wd}, \ref{fig:wd_imgnet} and \ref{fig:wd_topk_imgnet_supp}). We see that right before overfitting, the cosine match between \textit{per-epoch} curvature and FZ score is the same as the cosine match between \textit{cumulative} curvature and FZ score. The match starts to drop for per-epoch case and for the case without weight decay, as cross entropy maximization occurs at the later epochs, and far more aggressively without weight decay regularization. Hence, averaging over all the weights gives us the most stable and reliable results.
% \section{Quantitative Results}
% \label{sec:quan_res}

% % \subsection{Removal of high curvature samples}
% % In this section we compare the results of removing the high curvature examples to a similar proportion of randomly chosen examples from the dataset. We train on this reduced subset and report validation accuracy on the unaltered validation set. The reported numbers are the mean standard deviation over 5 runs.

% \clearpage
% \section{Discussion}
% \label{sec:disc}

% We wish to emphasize that we do not claim that curvature is the best way of finding mislabeled examples. Instead, we recommend that curvature should be one of the checks implemented, in conjunction with other checks, for a holistic view of dataset integrity. Our primary motivation with this manuscript is to show that curvature is a good measure of sample memorization, and can serve to find examples with conflicting labels and outliers or long tailed samples. 
\section{Conclusion}
\label{sec:conc}
Overparametrized networks are known to overfit to the training dataset, and often achieve 100\% training accuracy by memorizing training datapoints. \citet{zhang2017understanding} demonstrate an extreme version of this, wherein they show that the networks can fully memorize a training dataset with all labels randomized to perfect accuracy. This raises concerns about the networks consuming erroneous data, which can have an adverse impact when the decisions made by neural networks are used in real-life scenarios. In this paper, we propose curvature of the loss function around the datapoint, measured as the trace of the square of the Hessian, as a metric of memorization. We overfit to the training datasets, and measure the curvature of the loss around each sample. We validate curvature as a good measure of memorization in three ways. First, we visualize the highest curvature samples, and note that they are made of mislabeled, long-tailed, multiple class, or otherwise conflicting samples that are not clearly representative of their labels. Second, We also show that curvature estimates have a very high cosine similarity match with FZ scores, which are calculated extensively by training thousands of models per dataset. Instead, our method only requires training one network. Using our method, we catch a failure mode on CIFAR100 and ImageNet that is to the best of our knowledge, unobserved until now; that of duplicated images with different labels. Third, we show that using curvature to identify mislabeled samples in the case of synthetically mislabeled training sets achieves high AUROC scores. These three experiments help us establigh curvature as a reliable and scalable method for measuring memroization of a sample by a network. This can be utilized to check the integrity of different datasets, to identify areas that might be undersampled, or badly annotated. Finally, we study how curvature develops during training, helping us gain insight into the dynamics of overfitting.
\clearpage
\section*{Acknowledgement}
% \vspace{-1mm}
The authors wish to thank Dr. Andy Buchanan for the insightful discussions and suggestions. This work was supported in part by the Center for Brain Inspired Computing (C-BRIC), one of the six centers in JUMP, a Semiconductor Research Corporation (SRC) program sponsored by DARPA, by the Semiconductor Research Corporation, the National Science Foundation, Intel Corporation, the DoD Vannevar Bush Fellowship, and by the U.S. Army Research Laboratory.
% \bibliography{example_paper}
% \bibliographystyle{icml2023}

\bibliography{icml2023/example_paper}
\bibliographystyle{iclr2024_conference}

\appendix

%%%%%%%%%%%%%%%%%%%%%%%%%%%%%%%%%%%%%%%%%%%%%%%%%%%%%%%%%%%%%%%%%%%%%%%%%%%%%%%
%%%%%%%%%%%%%%%%%%%%%%%%%%%%%%%%%%%%%%%%%%%%%%%%%%%%%%%%%%%%%%%%%%%%%%%%%%%%%%%
% APPENDIX
%%%%%%%%%%%%%%%%%%%%%%%%%%%%%%%%%%%%%%%%%%%%%%%%%%%%%%%%%%%%%%%%%%%%%%%%%%%%%%%
%%%%%%%%%%%%%%%%%%%%%%%%%%%%%%%%%%%%%%%%%%%%%%%%%%%%%%%%%%%%%%%%%%%%%%%%%%%%%%%
\section{2D Toy Example Visualization}
\label{supp:2d}

We use a network of 7 fully connected layers of size [2,100,100,500,200,100,1000]. Each layer is followed by a batchnorm layer and ReLU layer. The network is trained for 150 epochs with an SGD optimizer and a learning rate of 0.1, momentum of 0.9 and a weight decay of 5e-4. There is only one minibatch per epoch. The training dataset consists of 15 points for each of the 2 classes, with a noise ratio of 0.3 introduced to 30\% of the data. The test dataset consists of 100 points for each class with no noise.
% We visualize an extended version of Figure \ref{fig:2d_viz} 
In Figure \ref{fig:logit_space}, at the bottom we show the input space, specifically, we visualize the probability heatmap of each point in the grid belonging to the red class. And we also add the visualization of the output space.
% \begin{figure*}[ht!]
%   \centering
%   \includegraphics[width=0.9\linewidth]{icml2023/pics/2d_viz.pdf}
%   \caption{Visualization of curvature of decision boundary around datapoints. Train and test datasets are shown on the top left, and loss curves on the top right. The bottom figure shows the decision boundary and input points in the input space and the logit space at different epochs of training.}
%   \label{fig:logit_space_appndx}
% \end{figure*}
Early epochs show large changes in the curvature around points in the input space. We can see from the output space that the focus of early epochs is to get all datapoints on the right side of the boundary. Once the datapoints are on the right side, the cross entropy loss still aims to maximize the log confidence of the correct class, by maximally separating the correct class logit from the others. In these epochs, while the logits get more separated, little change is seen in the input space. This leads to decreasing curvature after overfitting.

\section{Hyperparameters}
\label{supp:hyperparams}
Our estimator introduces two hyperparameters, $h$ and $n$. We tune h in the range of $\left[10^{-2},10^{-4}\right]$ by evaluating the cosine similarity between our scores and FZ scores. We note that input $X$ is in the range of $\left[0,1\right]$ (before mean-standard normalization) and all elements of $v \in \{+1,-1\}$. This means that the $L_2$ distance of the noise added to the input remains in the range of $||hv||_2 = h\sqrt D$, or equivalently, we add perturb each pixel by $\pm h$. Empirically, we found best results at $h=0.001$ and $n=10$. A larger $n$ gives very slightly better results, but increases compute cost.

\section{Network Architecture and Training details for Curvature Estimates}
We use modified versions of ResNet18 for all experiments with appropriately modified input sizes and channels. For MNIST and FashionMNIST, we use the downscaled ResNet \footnote{https://github.com/bearpaw/pytorch-classification/blob/master/models/cifar/resnet.py} with the average pooling layer downsized from 8 to 7 due to the reduced input resolution of MNIST and FashionMNIST. For CIFAR datasets we use the full-size ResNet18\footnote{https://github.com/kuangliu/pytorch-cifar/blob/master/models/resnet.py}. WE use PyTorch provided ResNet18 for ImageNet models. We use no augmentation for MNIST and FashionMNIST, and random horizontal flips and crops for CIFAR and ImageNet datasets. 
We train for 300 epochs on CIFAR datasets, with a learning rate of 0.1, scaled by 0.1 on the $150^{th}$ and $250^{th}$ epoch. For MNIST and FashionMNIST, we train for 200 epochs, with a learning rate of 0.1 scaled by 0.1 on the $80^{th}$ and $160^{th}$ epoch. For ImageNet we train for 200 epochs with a learning rate of 0.1, scaled by 0.1 on the $120^{th}$ and $160^{th}$ epoch. Where weight decay is used, its value is set to $10^{-4}$.
\label{supp:arch}

\section{Training Samples with Low and High Curvature}
\label{supp:train_viz}
\subsection{CIFAR10}

\begin{figure*}[ht!]
    \centering
    \includegraphics[width=1\linewidth]{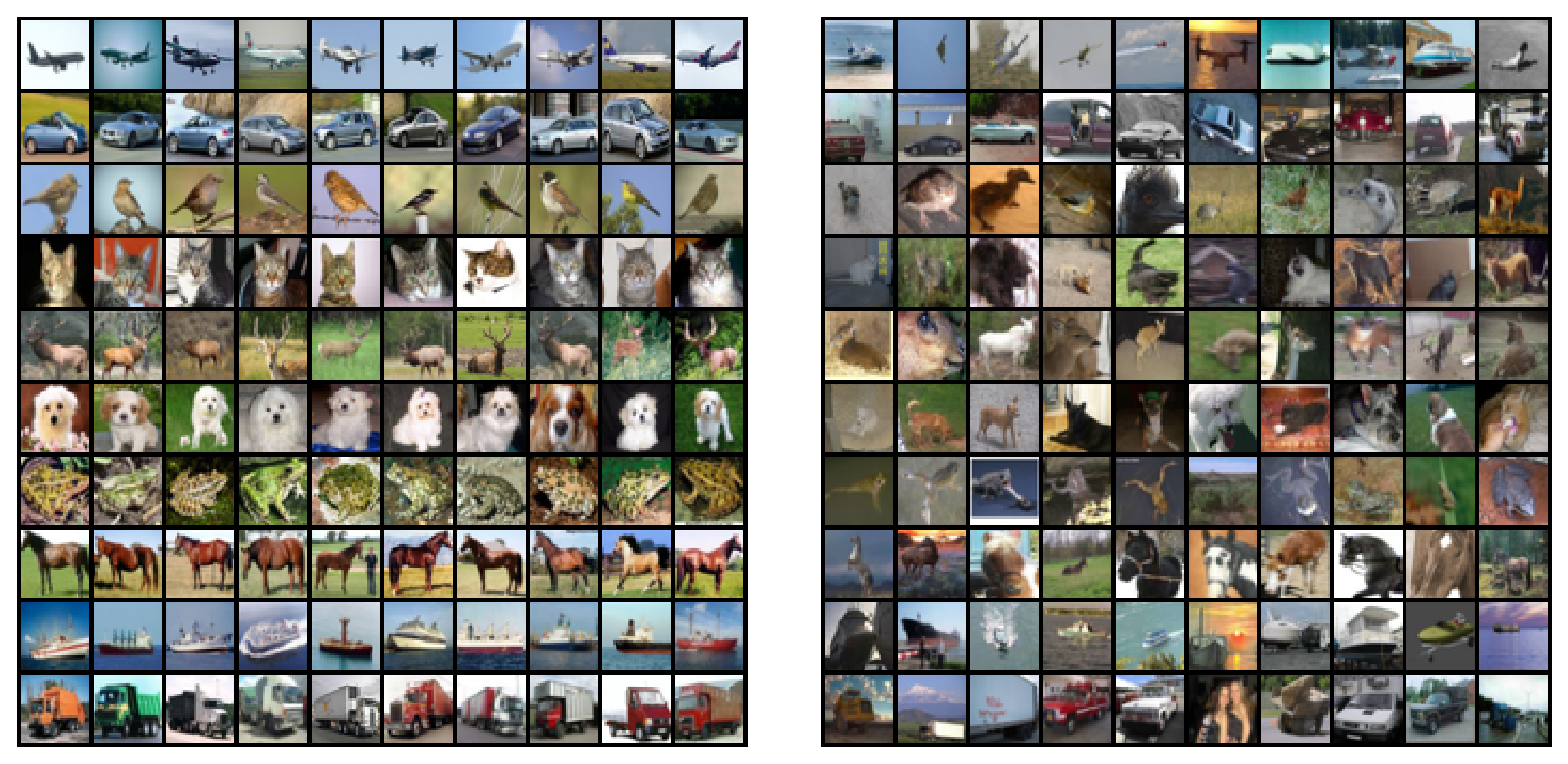}
    \caption{Low and High curvature samples from CIFAR10. Classes from top to bottom: [airplanes, cars, birds, cats, deer, dogs, frogs, horses, ships, and trucks]}
    \vspace{-3mm}
    \label{fig:cf10_train}
\end{figure*} 
The low curvature samples are shown on the left, with the high curvature samples on the right. The classes in order are: [airplanes, cars, birds, cats, deer, dogs, frogs, horses, ships, and trucks].
\newpage
\subsection{CIFAR100}
The low curvature samples are shown on the left, with the high curvature samples on the right. We show samples only from the first 10 classes, with the following labels:
[apple, aquarium fish, baby, bear, beaver, bed, bee, beetle, bicycle, bottle].

\begin{figure*}[!ht]
    \centering
    \includegraphics[width=1\linewidth]{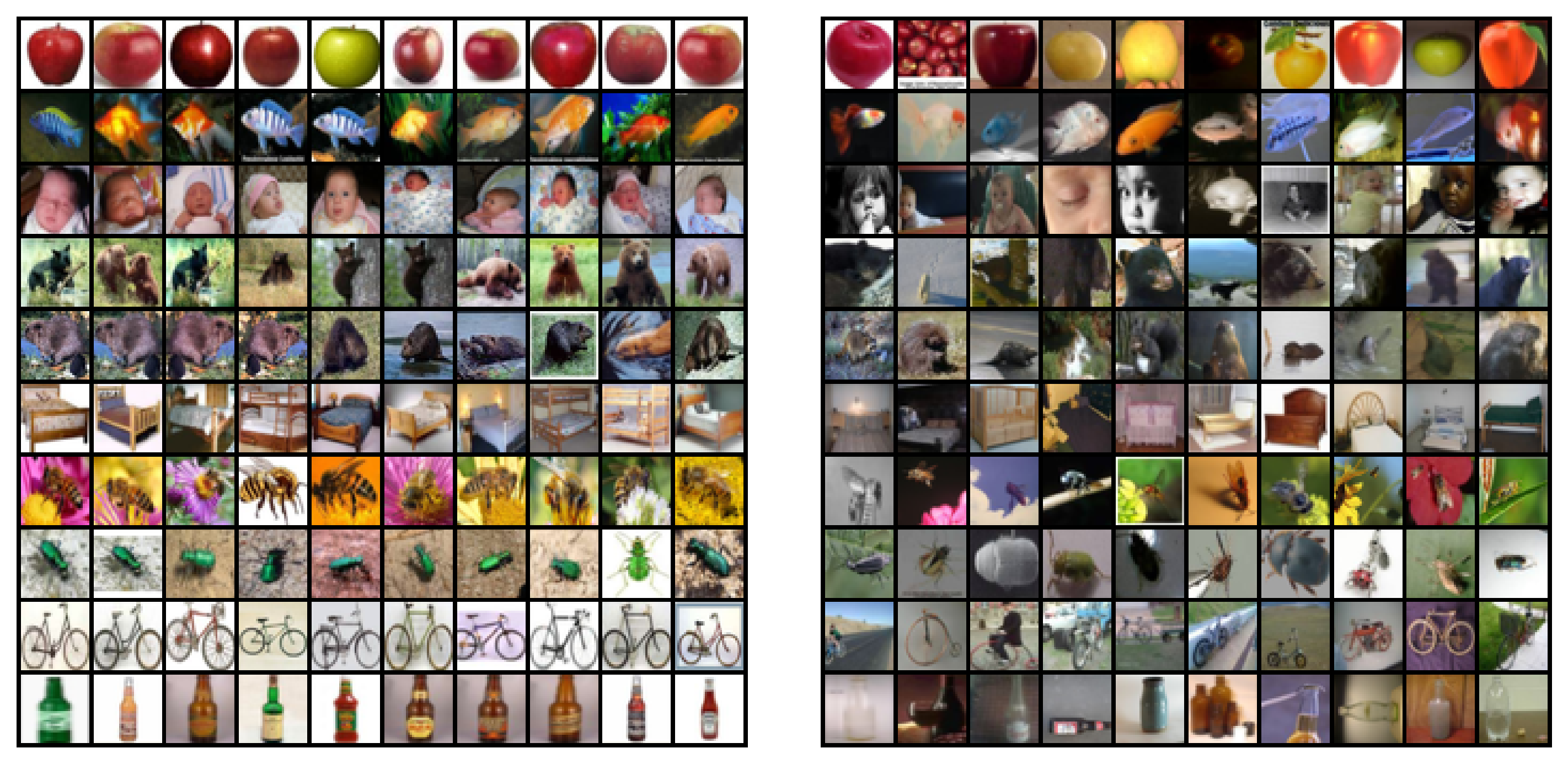}
    \caption{Low and High curvature samples from CIFAR100. Classes from top to bottom: [apple, aquarium fish, baby, bear, beaver, bed, bee, beetle, bicycle, bottle].}
    \vspace{-3mm}
    \label{fig:cf100_train}
\end{figure*}

\begin{figure}[ht!]
  \centering
  \includegraphics[width=1.0\linewidth]{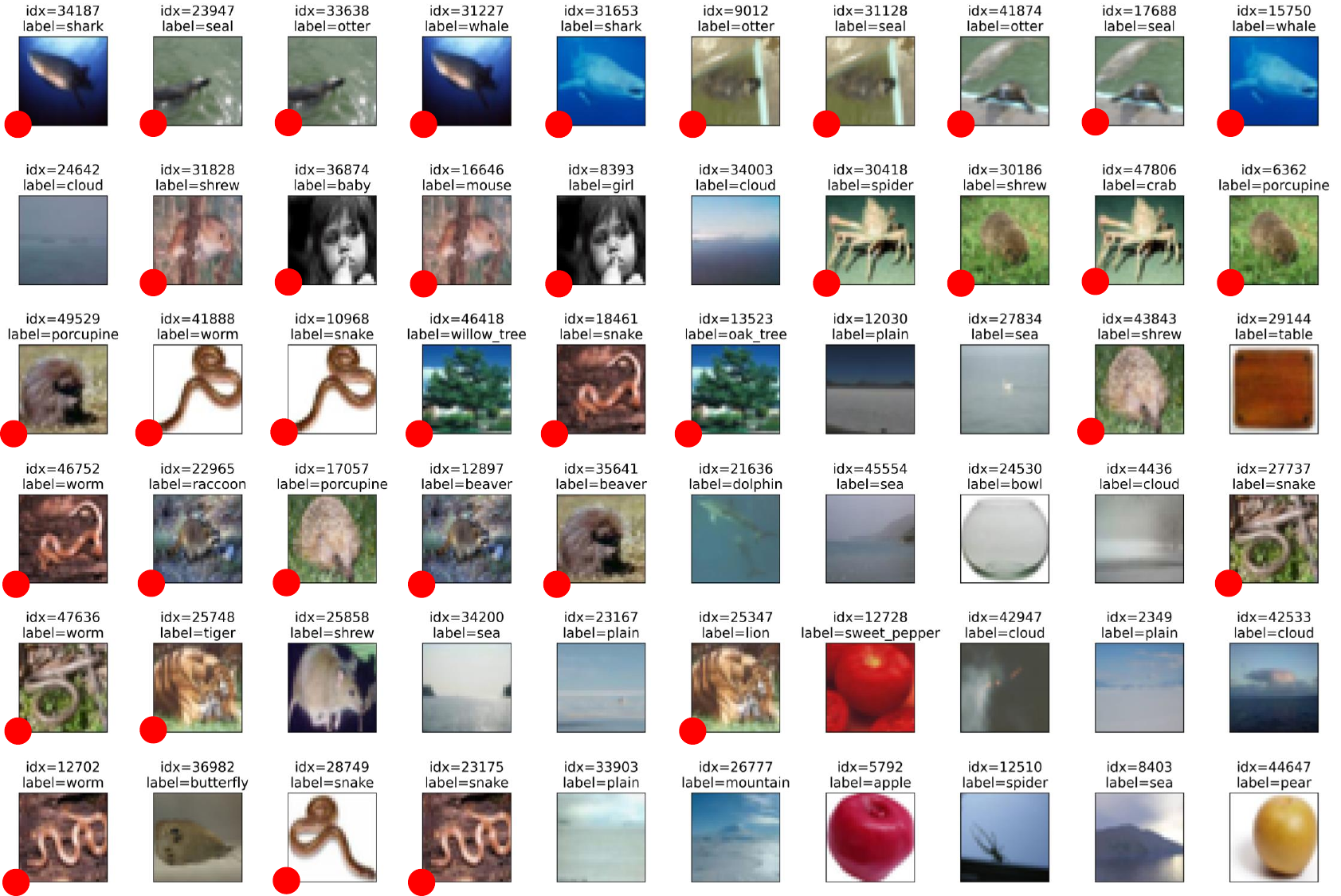}
    \caption{60 highest curvature samples from the training set of CIFAR100, identified by training on ResNet18. The index and label of the training sample are mentioned above the picture. We highlight samples that are duplicated with differing labels with a red dot.}
  \label{fig:cf100_pairs_appendix}
\end{figure}

\newpage
\subsection{ImageNet}
For ImageNet we show samples only from the first 10 classes, with the following labels:
[tench, goldfish, great white shark, tiger shark, hammerhead, electric ray, stingray, cock, hen, ostrich].
Note that low curvature samples on `tench' on ImageNet reveal a spurious correlation learnt by the network between tench and people holding the `tench' fish.

\begin{figure}[!ht]
\subfloat[Low curvature samples from ImageNet]{
\centering
\includegraphics[width=0.46\linewidth]{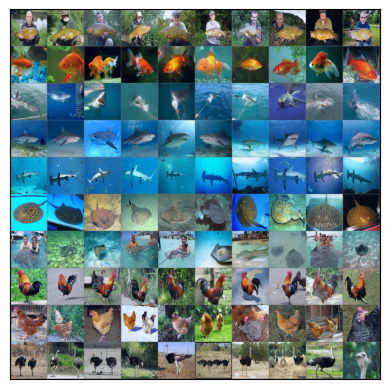}}
%no space
\hfill
\subfloat[High curvature samples from ImageNet]{
\centering
\includegraphics[width=0.46\linewidth]{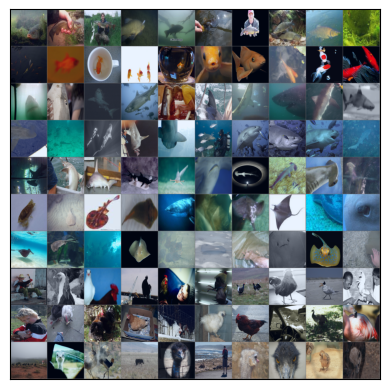}
}
\caption{Low and High curvature samples from ImageNet. Classes from top to bottom: [tench, goldfish, great white shark, tiger shark, hammerhead, electric ray, stingray, cock, hen, ostrich].}
%no space
\end{figure}

\begin{figure}[ht!]
  \centering
  \includegraphics[width=0.9\linewidth]{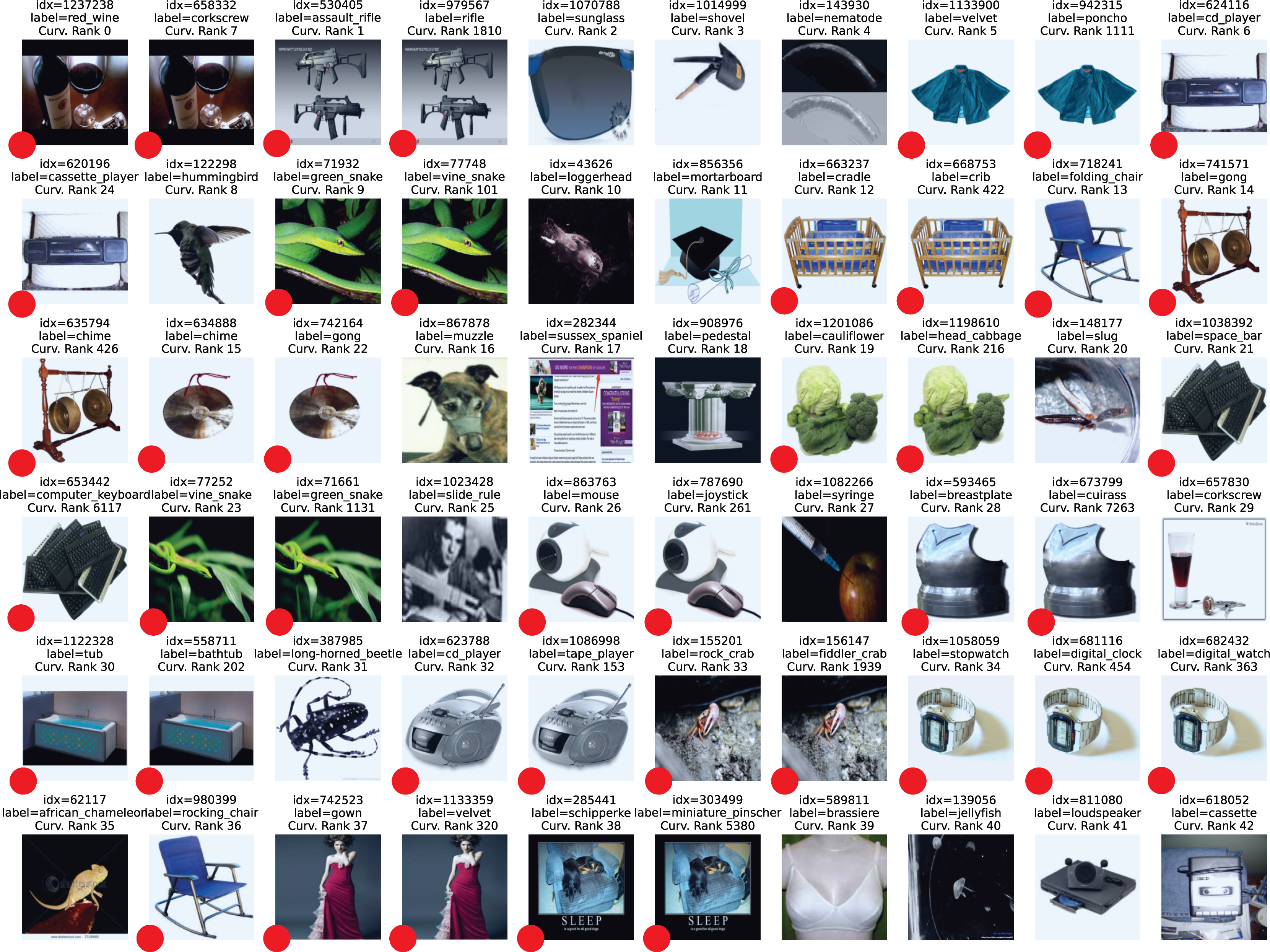}
 \caption{Top curvature samples from the training set of ImageNet along with duplicates with corresponding curvature ranks. Scores identified by training on ResNet18. We highlight samples that are duplicated with differing labels with a red dot.}
  \label{fig:imgnet_pairs_appedix}
\end{figure}

\subsection{Epoch-wise Cosine Similarity}
\label{sec:epoch_ise_appndx}
\begin{figure}[!htbp]
\centering
\subfloat[Cumulative and epoch-wise curvature of the training set of ImageNet with and without weight decay.]{\label{fig:wd_imgnet}
\centering
\includegraphics[width=0.48\linewidth]{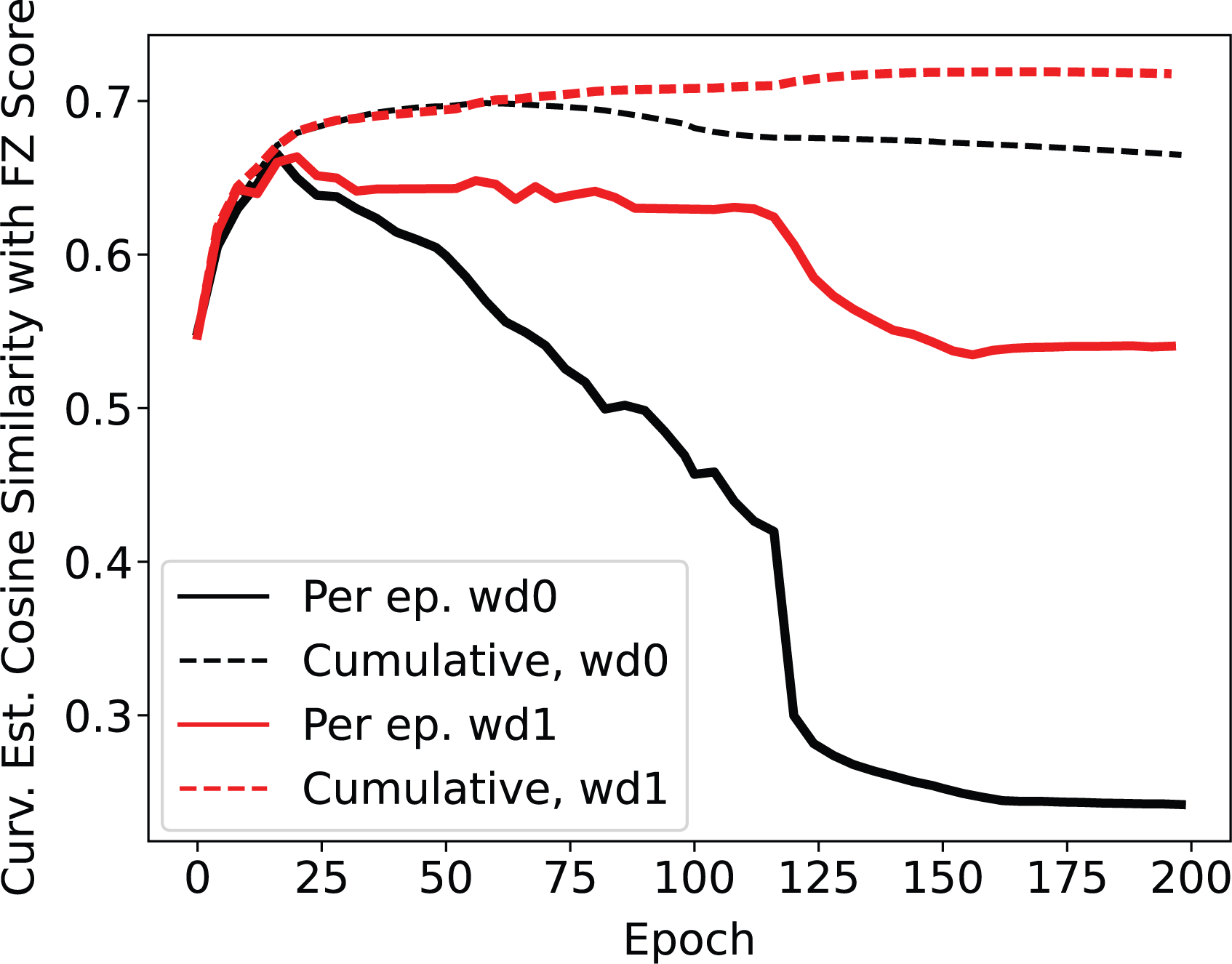}
}
%no space
\hfill
\subfloat[Cumulative and epoch-wise curvature of the training set of CIFAR100, with and without weight decay]{\label{fig:wd}
\centering
\includegraphics[width=0.46\linewidth]{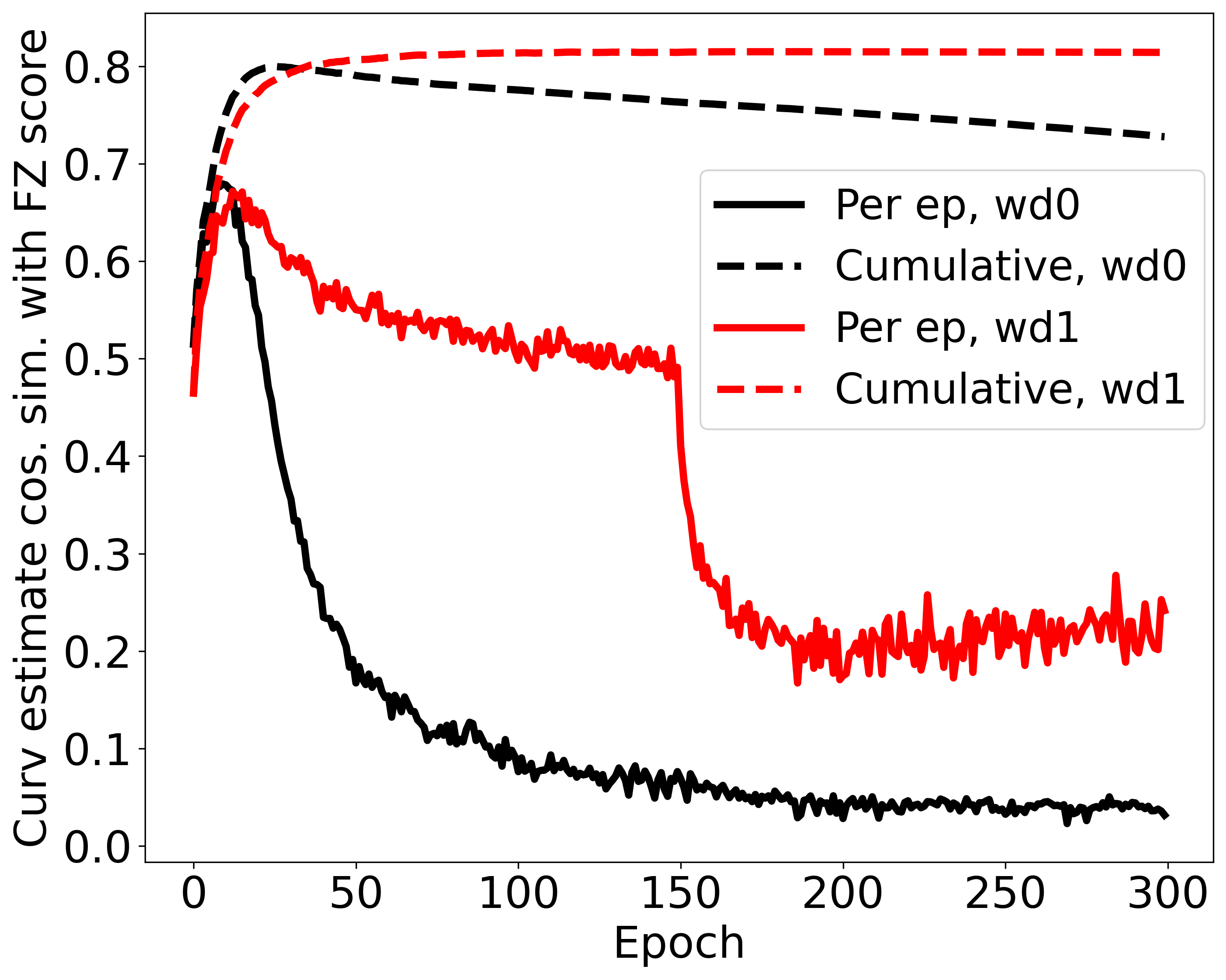}
}
%no space
\caption{Cosine similarity between curvature and FZ Memorization score for ResNet18 on ImageNet (left) and CIFAR100 (right). ImageNet results are plotted every 4 epochs for efficiency.}
\end{figure}

\begin{figure}[ht!]
\centering
\includegraphics[width=0.6\linewidth]{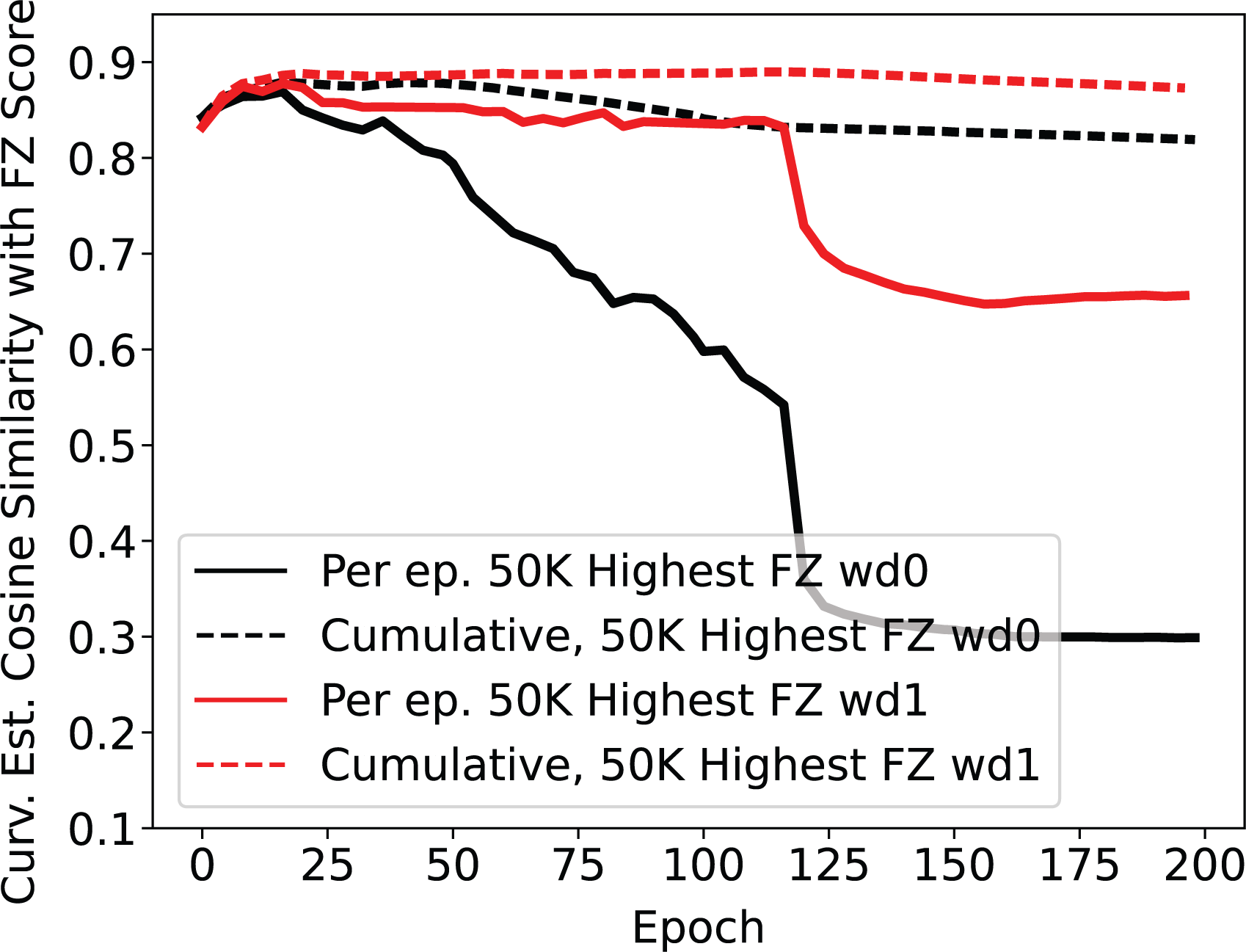}
\caption{Cumulative and epoch-wise curvature of the top 50K FZ samples of ImageNet with and without weight decay.}
\label{fig:wd_topk_imgnet_supp}

\end{figure}

We plot the epoch-wise and cumulative cosine similarity between the FZ score and curvature scores for CIFRAR100 (Figure \ref{fig:wd}) and ImageNet (Figure \ref{fig:wd_imgnet}). Further, we also plot the epoch-wise and cumulative cosine similarity between top 50K FZ sample memorization scores and the corresponding curvature scores for ImageNet (Figure \ref{fig:wd_topk_imgnet_supp}). We get very high similarity $\sim 0.9$

% \hfill
%     \subfloat[20 highest curvature samples from the training set of ImageNet. For extended figure see Figure \ref{fig:imgnet_pairs_appedix} in Appendix. ]{
%   \includegraphics[width=0.45\linewidth]{icml2023/pics/imagenet_high_curv_samples_main.png}
%   \label{fig:imgnet_pairs}
%   }
%   \label{fig:dataset_duplicates}
%   \caption{Highest curvature samples from the training set of and ImageNet (\ref{fig:imgnet_pairs}), identified by training on ResNet18. The index and label of the training sample are mentioned above in the picture. We highlight samples that are duplicated with differing labels with a red dot. Orange dots have duplicated pairs but are outside of the top 20.}
%   \end{figure}

\section{Validation Samples with High Curvature}
\label{supp:val_viz}
In this section, we train the same network as described in section \ref{sec:qual_res} on the validation sets of all 4 datasets and show the highest curvature samples for these. Results are of curvature averaged over all training epochs, and the network is trained without weight decay.

\begin{figure}[!ht]
\subfloat[MNIST]{
\centering
\includegraphics[width=0.45\linewidth]{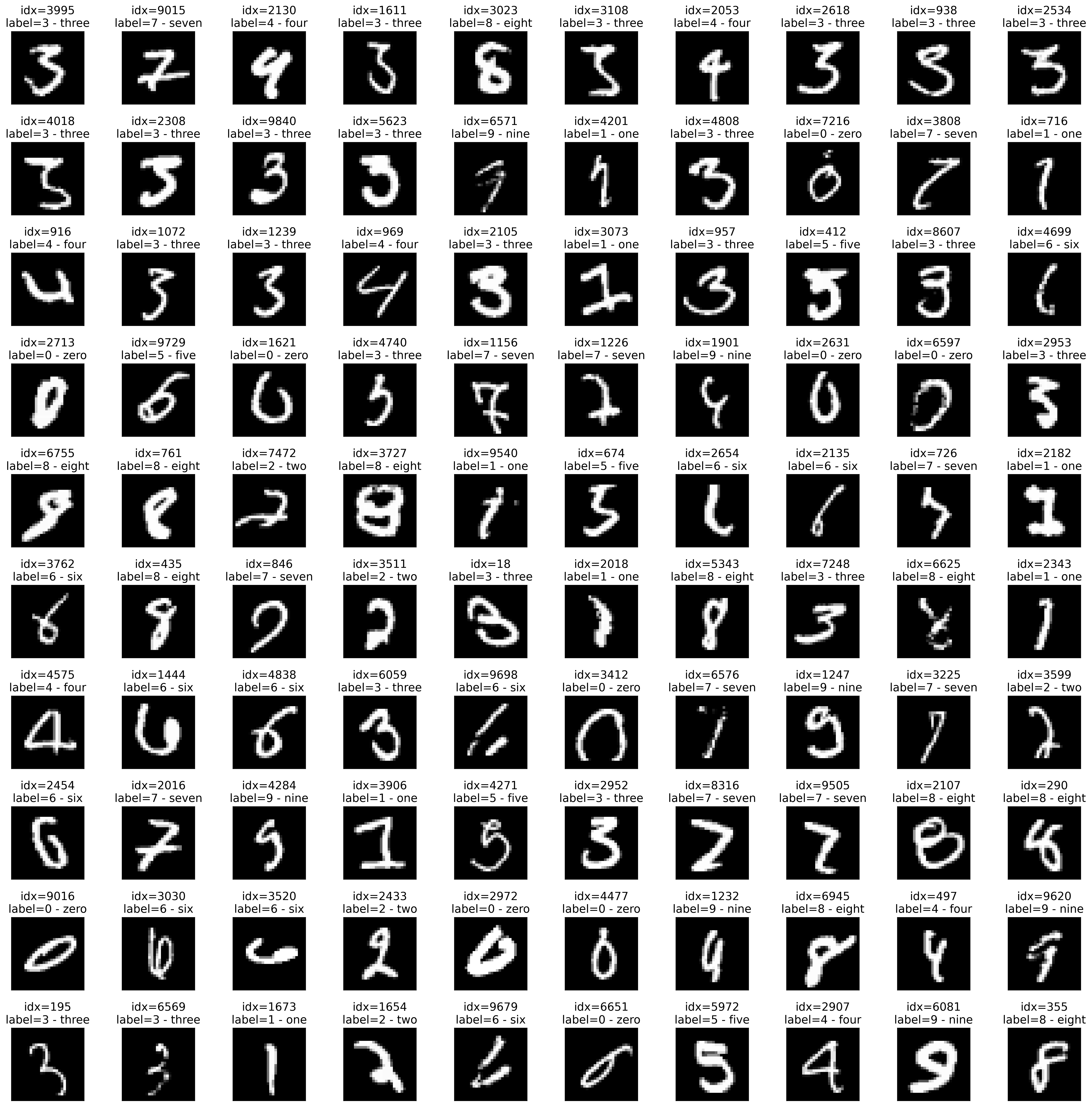}}
%no space
\hfill
\subfloat[FashionMNIST]{
\centering
\includegraphics[width=0.45\linewidth]{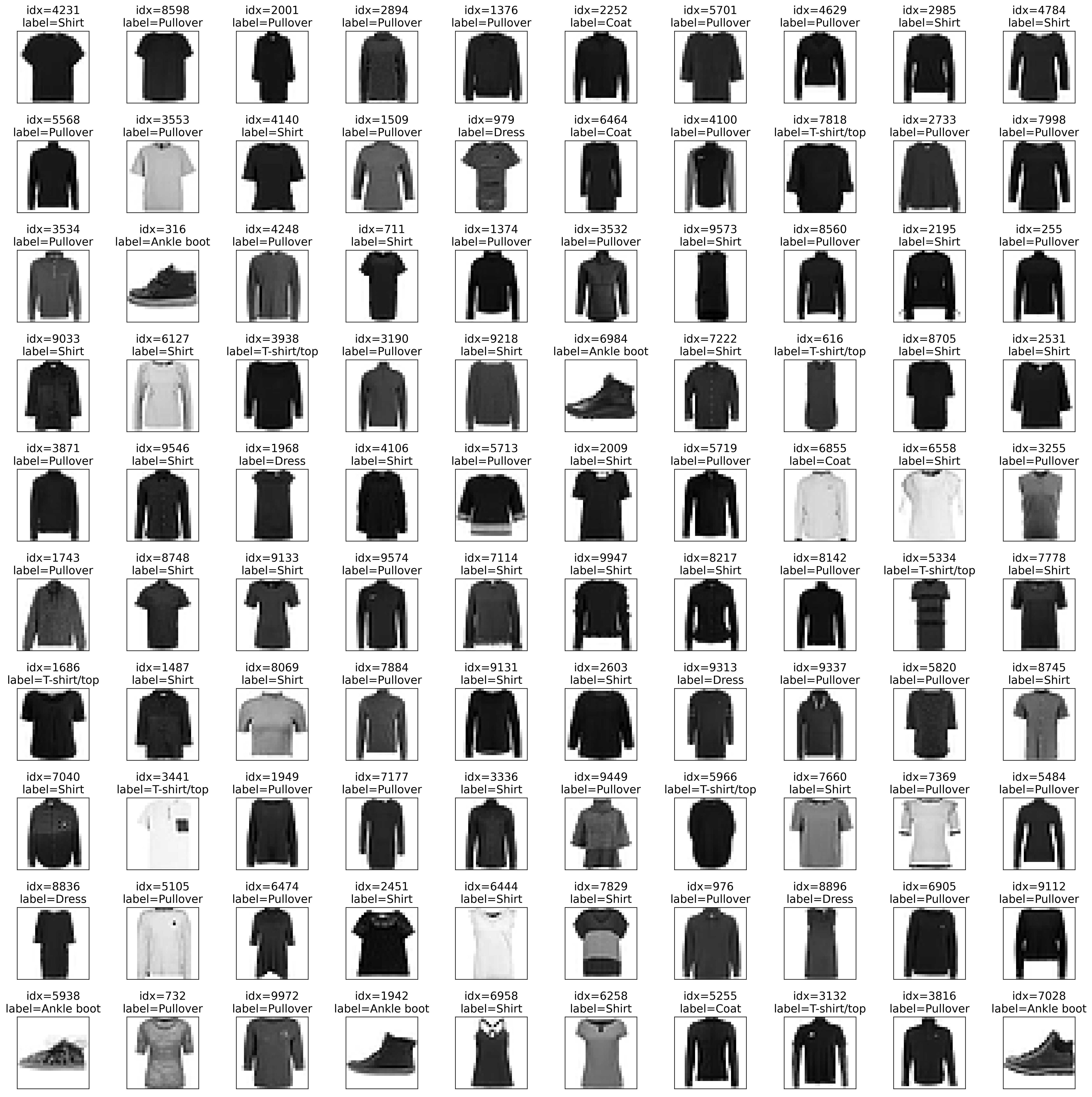}
}
\caption{High curvature samples from validation sets}
%no space
\end{figure}

% \begin{figure*}[!h]
%     \includegraphics[width=0.4\linewidth]{icml2023/pics/mnist_val.png}
%     \caption{High curvature samples from MNIST validation set}.
%     \includegraphics[width=0.4\linewidth]{icml2023/pics/fmnist_val.png}
%     \caption{High curvature samples from FashionMNIST validation set.}
% \end{figure*}

\begin{figure}[!ht]
\subfloat[CIFAR10]{
\centering
\includegraphics[width=0.45\linewidth]{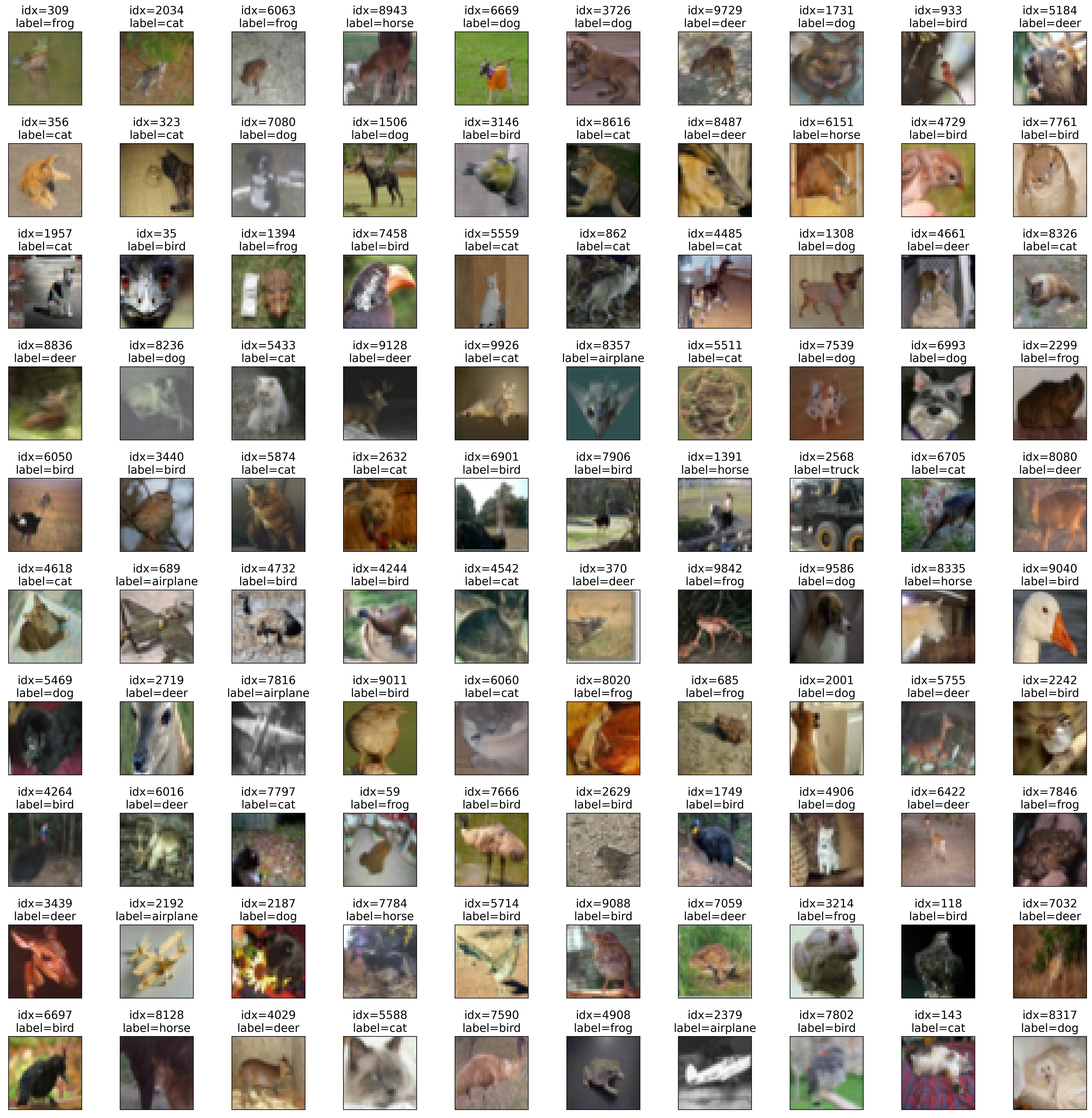}}
%no space
\hfill
\subfloat[CIFAR100]{
\centering
\includegraphics[width=0.45\linewidth]{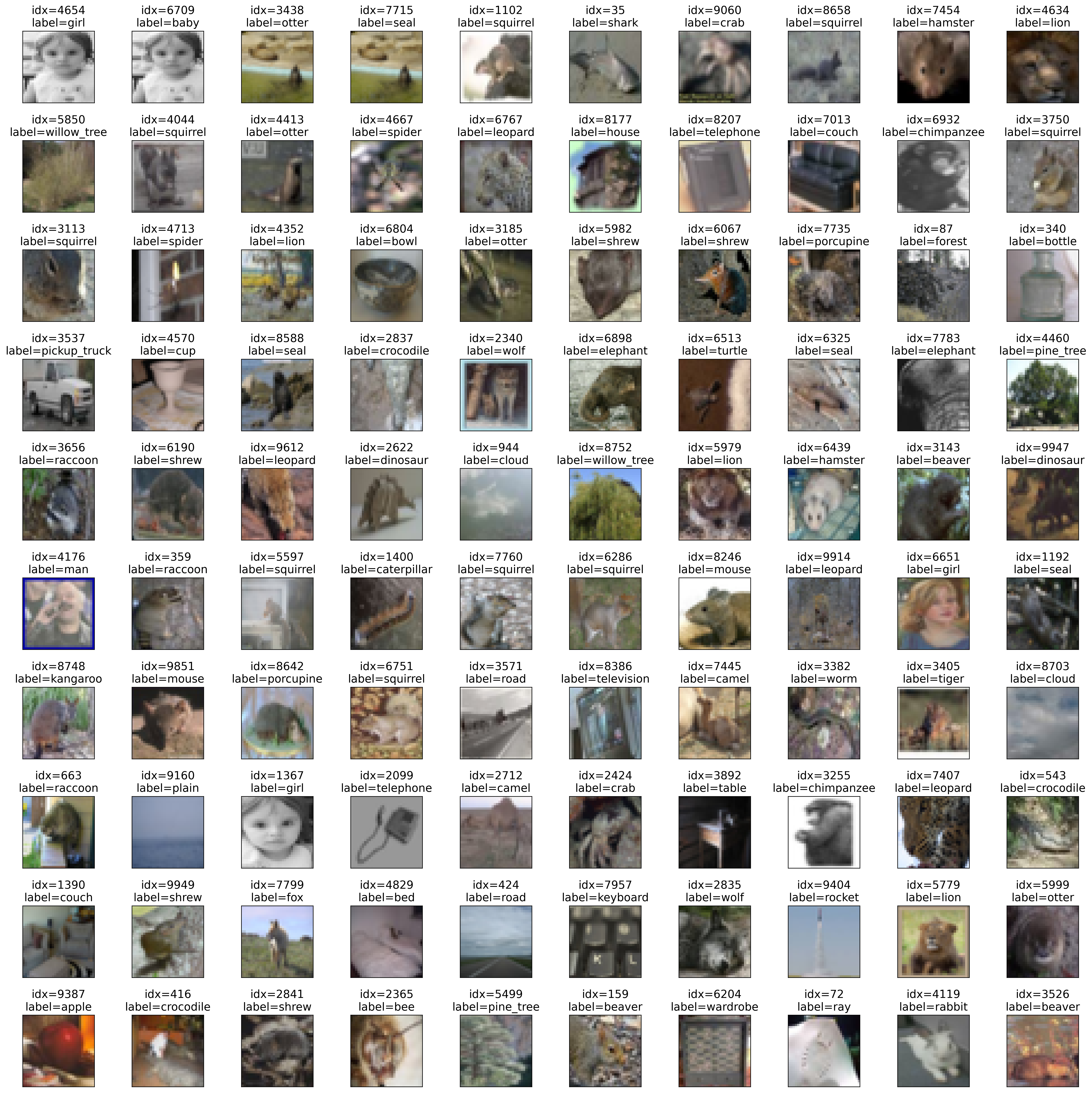}
}
\caption{High curvature samples from validation sets}
%no space
\end{figure}

\section{CIFAR100 most memorized samples}
\label{supp:cf100_worst}
Here we show the hundred most memorized examples as identified by FZ scores, and with curvature when training with weight decay $=1e-4$.
\begin{figure}[!ht]
\subfloat[Most Memorized according to FZ scores]{
\centering
\includegraphics[width=0.45\linewidth]{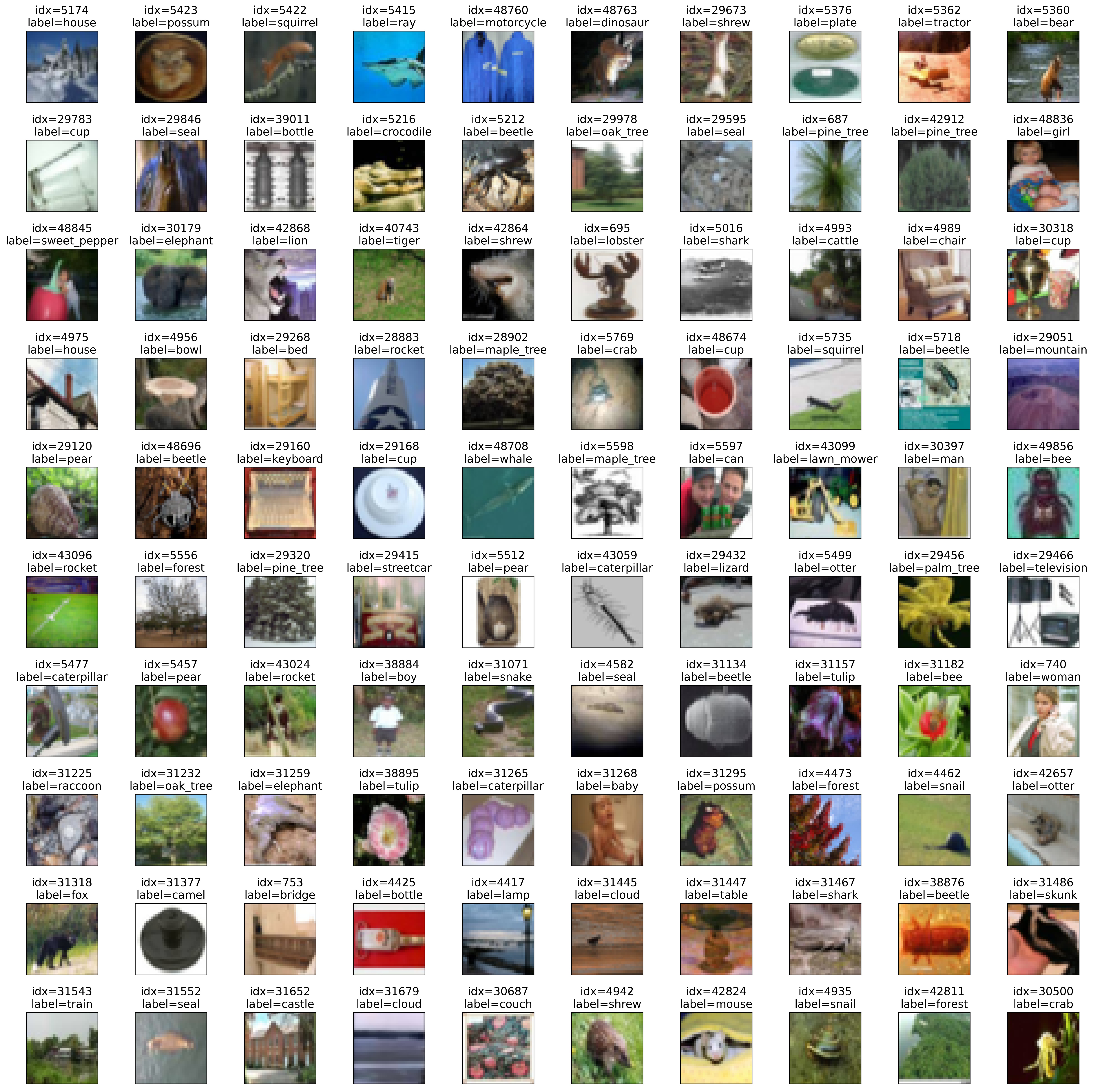}}
%no space
\hfill
\subfloat[Highest Curvature with Weight Decay on]{
\centering
\includegraphics[width=0.45\linewidth]{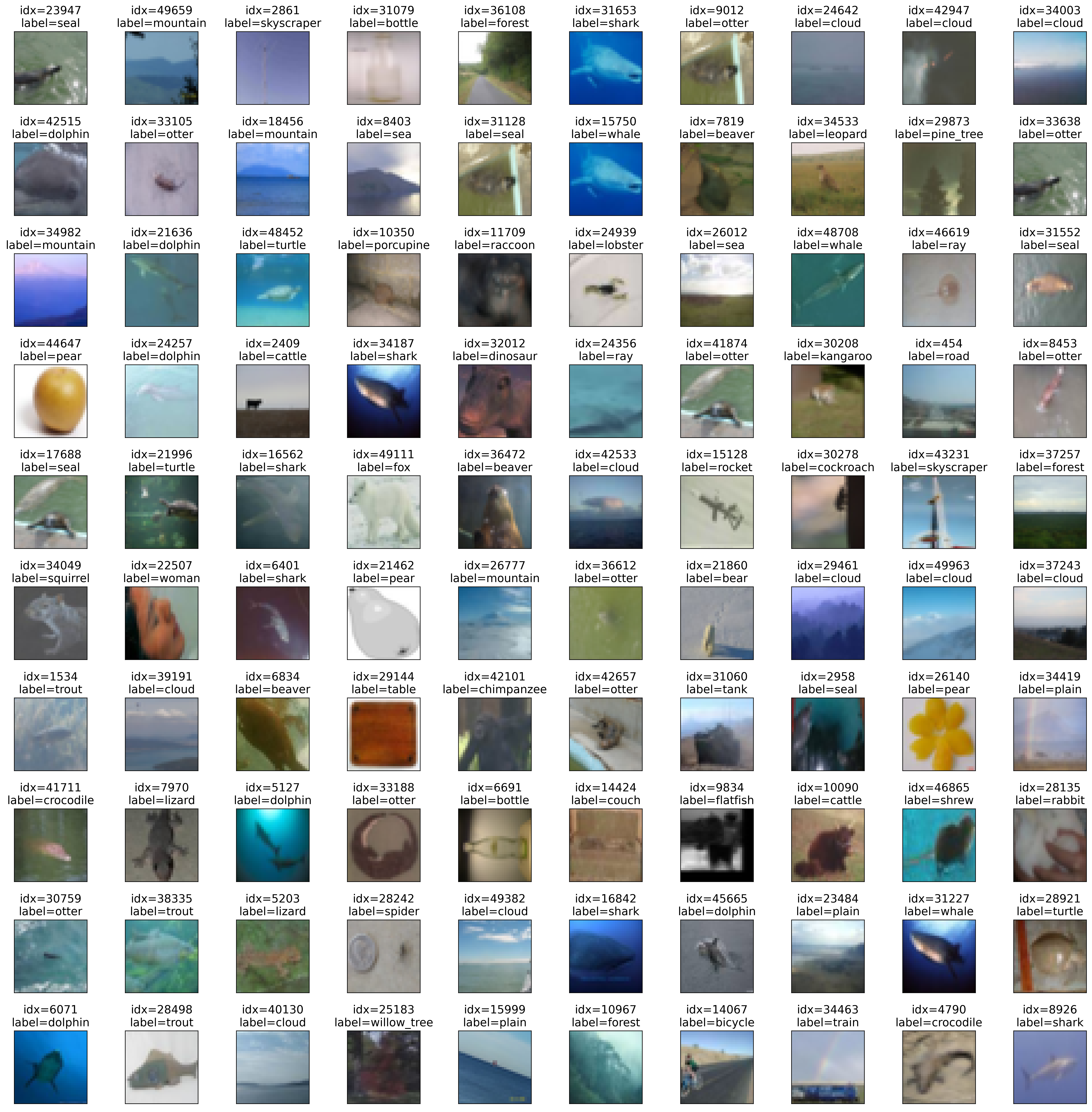}
}
\caption{High curvature samples from training sets of CIFAR100}
%no space
\end{figure}

% \section{CIFAR100 corruption results}
% \label{supp:cf100_corr}

%%%%%%%%%%%%%%%%%%%%%%%%%%%%%%%%%%%%%%%%%%%%%%%%%%%%%%%%%%%%%%%%%%%%%%%%%%%%%%%
%%%%%%%%%%%%%%%%%%%%%%%%%%%%%%%%%%%%%%%%%%%%%%%%%%%%%%%%%%%%%%%%%%%%%%%%%%%%%%%

\end{document}